\documentclass[12pt]{article}

\usepackage{scicite}
\usepackage{times} 

\usepackage{graphicx} 

\usepackage{upgreek} 
\usepackage{wasysym} 
\usepackage{titling} 
\usepackage{xcolor} 
\usepackage{ragged2e}
\usepackage{verbatim} 

\newenvironment{sciabstract}{%
\begin{quote} \bf}
{\end{quote}}

\title{Magneto-oscillatory localization for small-scale robots} 

\author
{F. Fischer$^{1,2}$ , C. Gletter$^{3,4}$, M. Jeong$^{3}$, T. Qiu$^{1,5,6,\ast}$\\
\\
\normalsize{$^{1}$Division of Smart Technologies for Tumor Therapy, German Cancer Research}\\
\normalsize{ Center (DKFZ) Site Dresden, Dresden, Germany}\\
\normalsize{$^{2}$Faculty of Engineering Sciences, University of Heidelberg, Heidelberg, Germany}\\
\normalsize{$^{3}$Cyber Valley Group – Biomedical Microsystems, Institute of Physical Chemistry, }\\
\normalsize{University of Stuttgart, Stuttgart, Germany}\\
\normalsize{$^{4}$Micro Nano and Molecular Systems Lab, Max Planck Institute for Medical}\\
\normalsize{Research, Heidelberg, Germany}\\
\normalsize{$^{5}$Faculty of Medicine Carl Gustav Carus, Dresden University of Technology,}\\
\normalsize{Dresden, Germany}\\
\normalsize{$^{6}$Faculty of Electrical and Computer Engineering, Dresden University of Technology,}\\
\normalsize{Dresden, Germany}\\
\\
\normalsize{$^\ast$E-mail: tian.qiu@dkfz.de}
}

\date{}

\begin{document} 

\baselineskip24pt 

\maketitle  

\begin{sciabstract} 

Magnetism is widely used for the wireless localization and actuation of robots and devices for medical procedures. However, current static magnetic localization methods suffer from large required magnets and are limited to only five degrees of freedom due to a fundamental constraint of the rotational symmetry around the magnetic axis.
We present the small-scale magneto-oscillatory localization (SMOL) method, which is capable of wirelessly localizing a millimeter-scale tracker with full six degrees of freedom in deep biological tissues.
The SMOL device uses the temporal oscillation of a mechanically resonant cantilever with a magnetic dipole to break the rotational symmetry, and exploits the frequency-response to achieve a high signal-to-noise ratio with sub-millimeter accuracy over a large distance of up to 12 centimeters and quasi-continuous refresh rates up to 200 Hz.
Integration into real-time closed-loop controlled robots and minimally-invasive surgical tools are demonstrated to reveal the vast potential of the SMOL method.

\end{sciabstract}

\section*{Introduction} 

Small-scale robots hold enormous potential for minimally-invasive medicine, for example, they may allow targeted drug delivery, \textit{in vivo} sensing, and new minimally-invasive surgical procedures \cite{nelson2010,schmidt2020}.
Recently, significant progress in the field has been made to wirelessly power and actuate nano- to millimeter robots in  biological environments \cite{taylor2016,li2017}.
Controlled propulsion has been shown in \textit{in vivo} or \textit{ex vivo} experiments, such as microrobots swimming in mouse intestines using magnetic fields \cite{servant2015} and chemical fields \cite{gao2015,li2016}.
We demonstrated that magnetically-actuated helical nanorobots can propel through the vitreous humor of the eye for drug delivery to the retina \cite{wu2018} and it was shown that enzymatically active micropropellers can move through gastric mucin gels \cite{walker2015}.
Millimeter-sized robots were reported to move through various biological tissues for minimally-invasive surgery.
For example, screw-shaped magnetic robots are able to penetrate muscle tissue \cite{ishiyama2001,rahmer2018} and to navigate through vascular systems \cite{jeong2011}.
Dragging, tumbling and other means of actuation have also been reported to actuate small-scale robots for biomedical applications \cite{son2021,yang2020,bruns2020,jeong2023}.
Despite the progress in actuation, a commonly overlooked challenge for small-scale robotics is the real-time localization of the small robots in deep biological tissues.
Localization requires the detection of the robot’s position and its orientation, which are both crucial for feedback control of the robot in complex unknown environments inside the body and thus greatly affect the efficacy to perform biomedical tasks.

Localization of small-scale robots poses various technical challenges \cite{wang2018} that traditional medical imaging and device tracking techniques cannot fulfil.
First, to continuously control the locomotion, real time tracking is required, however, established medical imaging techniques, such as magnetic resonance imaging  (MRI), have limited refresh rates.
Radiation-based imaging methods, such as computer tomography (CT), positron emission tomography (PET) and fluoroscopy, are unsuitable for the continuous tracking of moving robots due to undesired harmful radiation.
Magnetic particle imaging (MPI) is a promising new imaging method for magnetic nanoparticles, which has been used for combined tracking and actuation with improved imaging rates \cite{gleich2005,nothnagel2016}. 
Second, intracorporeal localization of small-scale robots requires high spatial resolution as well as large tissue penetration depth.
Ultrasound (US) imaging was commonly applied as a real-time imaging method for such devices \cite{servant2015,yu2019,li2019,pane2022}.
However, intracorporeal ultrasonic waves suffer from multiple reflections and strong scattering that limit the imaging contrast and the localization resolution. Moreover, US imaging is often two-dimensional (2D), making it difficult to localize and track a moving robot in the 3D space.
In our previous work, we demonstrated that optical coherence imaging (OCT) can be used to localize nanorobots in the vitreous \cite{wu2018}, however, opacity of most biological tissues poses a common limitation to near-optical wavelengths.
Photoacoustic imaging \cite{wu2019,wrede2022} and fluorescent imaging \cite{yan2017} were also reported to track robots’ motion \textit{in vivo}, however, their penetration depth is limited.
Lastly, as small-scale robots aim for wireless functionality, tethered localization techniques, such as electromagnetic tracking systems with tethered magnetic sensors for endoscopes or catheters \cite{hummel2005,rivkin2021}, cannot be applied.

Wireless localization and tracking methods were developed for biomedical devices, such as endoscopic capsules and implantable tissue markers.
Electromagnetic (EM) fields were used to track prostate movement with multiple embedded coils to a spatial resolution below 2 mm and a temporal resolution of 10 Hz \cite{willoughby2006}, however, each coil is 8 mm long and at least two coils are required for localization.
Using a complex chip design, an integrated sensor circuit to measure the 3D magnetic field gradients to localize the device with a sub-millimeter resolution at 10 Hz was reported \cite{sharma2021}.
Magnetic fields, which are safe to use in the human body, were generated from embedded permanent magnets and measured by external sensor arrays to localize medical devices \cite{schlageter2001,son2016}.
However, the static permanent magnets need to be relatively large ($>$6 mm$^3$) in order to generate a measurable field in deep tissues ($>$5 cm) and are heavily influenced by nearby magnetic surgical tools \cite{nicolae2019}.
Overall, most mentioned trackers have a large size (at least one dimension surpasses 7 mm) making them unsuitable for the integration on small-scale robots.

Traditional localization method based on single permanent magnets can maximally achieve 5 degrees of freedom (DoF).
As the magnetic field of the device exhibits rotational symmetry around the axis of the magnetic moment, it is fundamentally impossible to measure the rotation angle of a robot around this axis.
However, the missing sixth DoF for a static magnet can play a crucial role for the feedback control.
For instance, magnetic helical robots often have a magnetic dipole moment perpendicular to the propulsion (helical) axis.
Without localizing the rotational DoF around the magnetic dipole, the pitch (or yaw) angle of the robot is missing.
Currently, their propulsion direction can only be interpolated by relating the current location to a previous location \cite{xu2022} or using additional onboard sensors \cite{popek2017,taddese2018}.
When the initial robot's propulsion axis is unknown, or when the movement of the robot is disturbed, the sixth DoF is lost, which leads to loss of control over the robot.
Another example is the capsule endoscope, where full 6 DoF information is very useful to improve the stitching of endoscopic images and relate them to a global reference frame \cite{iakovidis2015}.
Full 6 DoF localization has recently been achieved using a magneto-mechanical torsional resonator that consists of two opposing magnets \cite{philips2023}, however, the device does not exhibit a net magnetic moment in the far field and hence cannot be actuated by external magnetic fields.

Here, we demonstrate small-scale magneto-oscillatory localization (SMOL) with 6 DoF for magnetic robots to enable real-time closed-loop control.
SMOL combines the resonance properties of EM devices and the miniaturized footprint of a permanent magnet.
The tracker consists of a finite magnetic moment attached to a cantilever, which oscillates at a designed frequency near 100 Hz about a rotational axis perpendicular to the magnetic moment to break the rotational symmetry of the permanent magnet (Fig. 1).
The magnetic field generated by the 0.8 mm$^3$ oscillating magnet is measured by an array of 10 sensors and fitted to a magnetic field model by an optimization algorithm, so that all 3 translational DoF and 3 rotational DoF of the device are accurately determined.
We show that the SMOL tracker can be readily integrated in a miniature robot (Fig. 1A) or onto surgical tools, and the same magnet can be used for closed-loop localization and actuation under different magnetic field excitations in biological environments.
The results show that full 6 DoF tracking is achieved with an adjustable spatial and angular resolution from sub-millimeter to sub-100-micrometer for all translational DoF and down to sub-degree for all rotational DoF at distances up to 12 cm.
Real-time localization at around 5 Hz is achieved, with the possibility for an increase beyond 50 Hz using the segmentation of a single oscillation.  
The SMOL method requires simple instrumentation and exploits a unique frequency response of the device to maintain high signal-to-noise (SNR) ratios in a magnetically unshielded environment, thus it will open up promising opportunities to localize small-scale robots or medical tools deep inside the human body for clinical applications.

\section*{Results} 

\subsection*{Localization principle}

The SMOL method is based on the mechanical resonance of a cantilever structure with an attached finite magnetic moment $\mathbf{m}$, as shown in Fig. 1B.
The magnetic moment, \textit{i.e.} a permanent magnet, acts as both a magnetic actuator to excite the oscillation of the cantilever and as an emitter of a varying magnetic field for sensing.
A thorough derivation of the physical model and measured signal is presented in Supplementary Materials.
Upon application of an external magnetic field $\mathbf{B}_\mathrm{ext}$ perpendicular to the magnetic moment direction and cantilever direction, a torque $\tau$ (Eq. S1) acts on the magnet, resulting in a deflection of the cantilever by the angle $\uptheta$ up to the maximum angle $\uptheta_\mathrm{max}$.
Using an alternating current (AC) magnetic field at the resonance frequency $f_\mathrm{res}$ of the mechanical structure as excitation, a periodic deflection can be induced which decays as an underdamped harmonic oscillation (Eq. S2-S4). 
Depending on the rotation and translation of $\mathbf{m}$ in time, the magnetic field emitted from the magnet, approximated as ideal magnetic dipole (Eq. S5), produces a non-trivial signal (Eq. S10) of the magnetic flux density $\mathbf{B}$ with respect to $\uptheta$ at a nearby magnetic sensor at position $\mathbf{r}$.

The sensed magnetic field from a SMOL device exhibits, most dominantly, the fundamental frequency $f_\mathrm{res}$ of the mechanical oscillator but also multiples of this frequency (higher harmonics) due to the highly non-homogeneous shape of the dipole field.
The main components of this signal are the first and second harmonics due to the in-plane oscillation of the magnetic moment vector ($x$-$z$-plane in Fig. 1B).
By recording with multiple sensors simultaneously, \textit{i.e.} in a sensor array (Fig. 1A), the multitude of different signal shapes can be used to decode the 3D position and 3D orientation by fitting the physical model (Eq. S2 to S8) to recorded data.
Effectively, the oscillating magnet breaks the rotational symmetry of a static permanent magnet, resulting in full 6 DoF information of the SMOL device.

Three main components are necessary for a complete SMOL system, as shown in Fig. S1 (details in the Materials and Methods section).
First, an excitation unit, consisting of a set of coils, is driven at the known resonance frequency of the SMOL device.
Here, the excitation unit consists of two perpendicular flat coils to generate any planar B-field direction above the coils. 
Second, a sensing unit in the form of a magnetic sensor array picks up AC magnetic signals of the SMOL device. Here, ten fluxgate magnetometers with sub-nanotesla resolution are arranged in a 15 $\times$ 15 cm$^2$ plane.  
Third, the SMOL device itself, shown in Fig. 1A, is the key component and the device to be tracked.
Its mechanical resonance frequency is tuneable by change of cantilever properties, such as elasticity or dimensions of the cantilever and the magnetic mass, which is crucial to achieve spectral separation from magnetic noise frequencies.
For the following characterization, a steel cantilever (length $\sim$3.5 mm, width 0.2 mm, thickness 20 µm) and a N52 NdFeB magnet of dimensions $\diameter$ 1 mm $\times$ 1 mm are used (see also Materials and Methods).
In order to avoid rapid damping of the oscillatory motion in direct contact to the environment, an enclosure surrounds the cantilever and magnet.
Besides the protective function, it also limits the maximum deflection angle $\uptheta_\mathrm{max}$, which is of importance for a precise localization.

\subsection*{Oscillation of a SMOL device}

A high-speed oscillation of a SMOL device with $f_\mathrm{res} = 135$ Hz is presented in Movie S1 and shown at two frames with the highest deflection angles of approximately -20° to +20° in Fig. 1C.
The corresponding image analysis is shown in Fig. S2, while a schematic representation of a full time sequence of a single measurement is shown in Fig. 1D.
The measurement can be divided into an excitation phase and a signal phase.
In the excitation phase, a current $I_\mathrm{coil}$ in an unbiased sine-wave form is applied to the excitation coil at the  oscillating system's resonance frequency.
Multiple cycles are used to gradually increase the deflection angle $\uptheta$ to the maximum angle.
Since the excitation field's magnitude is beyond the measurement range of the magnetic sensors, the sensors $B_1$ to $B_n$ saturate periodically during the excitation phase.
In the signal phase, the excitation field is shut off and the stored energy in the cantilever is released in an underdamped oscillatory movement.
During this period, $\uptheta$ gradually decreases and meanwhile the sensors pick up the magnetic signal.

A recorded raw signal is presented in Fig. 1E.
The presence of magnetic noise in combination with a highly complex magnetic signal from the SMOL device lead to the necessity of signal filtering.
The full signal filtering process, together with discrete Fourier transformations (DFT), is presented in Fig. S3 and further explained in Supplementary Materials.
Using a spatial difference, low-pass filtering and a time derivative, high fidelity reconstruction of the oscillating signal (Fig. 1F) with a 20-fold increase of the signal-to-noise ratio (SNR, with respect to 50 Hz noise) is achieved.
Regarding only the first signal periods, the SNR amounts to approximately 61.

All necessary signal features for localization can be extracted by utilizing an integer multiple number $N$ of a half period of a filtered signal, \textit{i.e.} a half period of the magnets' oscillation.
In Fig. 1G, a full period ($N = 2$) is shown, which lasts 10 ms and the corresponding magnet deflection is illustrated below.
Per half period, the signal is down-sampled to $n = 4 N +1 = 5$ equidistant discrete points, which allow a detection up to the fourth harmonic for the 8 non-redundant points per full period, according to the Nyquist-Shannon sampling theorem.
These data points, for all magnetic sensors, and their respective times are fed into an unweighted Levenberg-Marquardt optimization algorithm \cite{more1978} using Eqs. S2-S8 as a physical model with the 3D position $x,y,z$ and 3D orientation, given in quaternions $q_1$ to $q_4$, as fitting parameters.
The sum squared error SSE (Eq. S20) between real data and the model fit is minimized during the process and coefficients of determination R² (Eq. S23) of over 0.99 are achieved, indicating a satisfactory fitting results while also validating the physical model.
Since the localization result is dependent on good estimations of $\uptheta_\mathrm{max}$ and damping coefficient $\upeta$, a calibration is performed when first using or changing the physical boundaries of the SMOL device by adding both variables as optimization parameters and fixing the $z$-component to the true value (see Supplementary Materials).

\subsection*{Spatial and angular accuracy}

Experiments and numerical simulations were carried out to determine and validate the localization accuracy of the SMOL method.
Since a visual ground truth with respect to the measurement plane ($x$-$y$-plane) is difficult to establish with an accuracy below 1 mm, differential measurements were performed.

Figs. 2A-C show the localization accuracy for two translation directions, $x$ and $z$, and the rotation axis $\phi_z$, which cannot be determined for a static magnet, \textit{i.e.} the intrinsic $z$-axis (see Fig. 1B).
A summary table of all 6 DoF accuracies is given in Fig. 2D.
The localization accuracy is distinguished for two modes: The speed mode uses two half-periods ($N = 2$) for faster localization rates owing to the low amount of data for optimization, whereas the precision mode (shown as inset) uses twenty half-periods ($N = 20$) for a more precise localization but lower refresh rates.
The complete translation and rotation measurements are presented in Figs. S4 and S5, respectively.

For the speed mode, over the large investigated in-plane translation ranges in $x$- and $y$-direction at 80 mm distance from the sensing array, as well as for translation in $z$-direction up to 120 mm distance, sub-millimeter accuracy is achieved, indicated by the perfect match of values to the 45 degree ground truth line.
For the precision mode, measured in 200 µm steps, accuracies significantly below 100 µm are achieved, which is less than one-tenth of the magnet's length at a distance of approximately 100 times of the magnet's length. 
In $z$-direction (Fig. 2B), an increasing scattering of the points with increasing distance is observed, which suggests a decrease in the measurement precision at large depths.
The main reason is the magnetic field decay over the cube of the distance according to Eq. S5.

Full 360° rotations are performed for all three rotation axes in speed mode.
Accuracies below 4° are achieved for all axis, whereby the intrinsic $z$- and $y$-axis perform, on average, slightly better than the $x$-axis.
This discrepancy, which is also prominent in simulation, is likely caused by a minor amplitude decrease from low-pass filtering or direction-dependent noise.
In precision mode, measured in increments of 1°, accuracies below 0.5° are achieved for all axes.

Simulation results are also shown in Fig. 2A-D and match very well to the experimental results for all accuracy measurements.
It reveals that the numerical model (see Supplementary Materials) clearly represents important features of the SMOL method, as it considers real magnetic noise in all directions.
The agreement between simulation and experiments validates the numerical model, and thus the latter offers a valuable way to predict and optimize the performance of SMOL devices.
A detailed explanation of accuracy measurements and the underlying statistical analysis (Eqs. S24-S30) can be found in Supplementary Materials.

Using the numerical model, SMOL devices with a scaled cubic magnet of side length $a$ (volume $V = a^3$) are simulated and shown in Fig. 2E, while keeping the resonance frequency, deflection angle and damping constant.
The localization depth where sub-millimeter precision (mean standard deviation in all directions $\sigma_{x,y,z} < 1$ mm) can still be achieved, defined as $z_\mathrm{max}$, is between 110 mm and 115 mm for the experimentally characterized SMOL device.
A clear linear trend between $z_\mathrm{max}$ and $a$ can be observed which can be ascribed to the scaling behavior of B-field according to Eqs. S5 and S8 where $B \sim V/z^3 \sim (a/z)^3$ for a SMOL device in the center of the array.
At large distances $>$100 mm, the sensor array is additionally scaled proportional to the magnet size to compensate the diminishing signal difference of adjacent sensors.
An optimized, realistic SMOL device achieves a localization depth of 156 mm (details in Supplementary Materials).

\subsection*{Precision and localization rate}

The highly tuneable range of the precision, given as standard deviation, across all directions $\sigma_{x,y,z}$ at $z = 80$ mm, in addition to the localization rate $f_\mathrm{loc}$ (including excitation and coil ringdown time), is analyzed for different numbers of evaluated half periods $N$ in Fig. 2F.
With increasing $N$, meaning that a larger portion of the signal is evaluated, $\sigma_{x,y,z}$ decreases from 0.5 mm for $N = 1$ to 0.1 mm for $N = 20$.
The decay can be attributed to an averaging effect and is most significant between $N = 1$ and $N = 6$.
Random noise or noise at frequencies distant from the resonance frequency of the SMOL signal is simply averaged out, as it does not reoccur with the same magnitude and in phase with the signal over multiple half-periods. 
In contrast, noise, appearing in very close spectral proximity to the resonance frequency, adds an unknown contribution to the signal in phase with the SMOL signal, creating an uncorrectable error in all half-periods which amounts here to approximately 0.1 mm.
While $N$ is increased, the localization rate automatically decreases due to a higher number of data points $n$.
For the presented system, the total number of points, which are fed into the optimization algorithm, for 10 sensor signals with 5 sampling points each for $N $ is 50 ($10\times(4N +1)$), whereas for $N = 20$ it is 810.
Correspondingly, with the current system,  $f_\mathrm{loc}$ of up to 8 Hz can be achieved for the {\it{low}} precision setting ($\sigma_{x,y,z}$ $<$ 1 mm) and 2.5 Hz for the {\it{high}} precision setting ($\sigma_{x,y,z}$ $<$ 200 µm).
Computation times required per localization are currently 40-60 ms for the speed mode and $\sim$360 ms for the precision mode.
Overall, the upper limit of the localization rate is mostly defined by the hardware system and could be further reduced using, for example, single-pulse excitation, better current amplifiers and higher computational power.
The best trade-off between precision and localization rate is found between $N = 4$ and 6. 

Since SMOL is based on the mechanical response of the cantilever, the precision is influenced by the mechanical coupling between the outside environment to the housing \cite{fischer2022}. 
In Fig. 2G, a wide damping spectrum is simulated, ranging from no damping ($\upeta = 0$) to very strong damping ($\upeta > 25$ s$^{-1}$) assuming exponential damping (Eq. S4).
For reference, experimentally determined damping coefficients for a solid boundary and a viscous liquid boundary (Glycerol) are 1.6 s$^{-1}$ and 31.7 s$^{-1}$, respectively, highlighting the extremely diverse environments in which SMOL is generally applicable.
A linear increase of $\sigma_{x,y,z}$ in dependence of $\upeta$ can be observed for both speed mode and precision mode, the respective slopes of a linear regression fit are 10.3 µms (R² = 0.90) and 6.8 µms (R² = 0.92).
This means that the SNR is better for a SMOL device in more rigid environments.
Even though the signal is quickly decayed to 5\% of the original magnitude after the first 120 ms of a signal for high damping of $\upeta = 25$ s$^{-1}$, in precision mode $\sigma_{x,y,z}$ is still very high with $\sim$200 µm.
Beyond this damping coefficient, $N$ has to be reduced to obtain a sufficient SNR for localization. 

\subsection*{Superfast localization}
Excitation and coil ringdown times, which are contributing adversely to $f_\mathrm{loc}$, are both inherently limiting the maximally achievable rates.
Significantly higher $f_\mathrm{loc}$ can be obtained by directly segmenting a signal into smaller partitions, each with the same number of half periods $N_{\mathrm{seg}}$.
Reducing $N_{\mathrm{seg}}$ to 1 means that every half period is independently evaluated, resulting in a {\it{superfast}} localization rate limit $f_\mathrm{loc,limit}$ of twice the resonance frequency. 
This means that a SMOL device with $f_{\mathrm{res}} = 103.5$ Hz can be temporarily localized with a 207 Hz refresh rate, which is beyond the necessary rate for real-time medical applications.
Figs. 2H and S6 demonstrate the capabilities of this method to localize a fast linear stepping motion with a speed up to 200 mm/s.
High accuracies are already obtained with a refresh rate of 51.7 Hz for a total duration of 2.4 s (Fig. 2H) and the precision is better than 1 mm in the first 1.5 s of the signal (Fig. S6F), and then it requires re-excitation.
The re-excitation of the SMOL oscillation takes approximately 80 ms, for both the excitation and the coil ringdown with the current system, which corresponds to the total downtime of the localization.
Therefore, a quasi-continuous localization with superfast rates can be achieved. 
The {\it{superfast}} localization approach pushes the SMOL method to the physical limit and can be used for tracking of very fast movements, given the necessary computation speed and low damping of the oscillation.

\subsection*{Integration of SMOL and actuation in millirobots}

The single magnet in the SMOL device can also be used for the actuation of small-scale robots \cite{yang2020}.
However, since the magnet is fixed to a delicate elastic component, special care has to be taken when combining it with the large forces needed for magnetic actuation.
Two kinds of millirobots, as shown in Fig. 3, demonstrate not only the compatibility of SMOL with magnetic actuation but also its versatility of implementation.

Fig. 3A presents a magnetic gradient force actuated millirobot being dragged through a viscous environment in an {\it{R}}-shaped path. 
This SMOL robot has a magnetic moment axis aligned with the cantilever, resulting in a pulling force $F$ upon application of an appropriate magnetic gradient field.
The force is applied away from the cantilever to avoid damage from sudden compression.
External B-fields are generated using four independently controlled electromagnetic coils (Fig. S8A) for arbitrary in-plane movements.
SMOL and actuation are performed sequentially with an average refresh rate of 3.5 Hz, enabling real-time closed-loop control of a wireless magnetic robot without any optical feedback.
The optical path aligns very well to the SMOL path, as also shown in Movie S2 and analyzed in Fig. S9, with a tracking error below 0.3 mm. 

Fig. 3B presents a torque-actuated helical millirobot penetrating a viscoelastic gel in an {\it{S}}-shaped path.
During actuation, a magnetic torque around the cantilever axis is applied to the magnet and the cantilever is able to transfer the torsion to the housing, which leads to rotation of the SMOL device.
The robot has a helical shape on the surface to couple the rotation to the translation, and thus propels in soft viscoelastic materials \cite{ishiyama2001,li2019}.
The actuation of the helical millirobot is performed wirelessly using an external rotating magnetic field generated by a rotating permanent magnet with a steerable rotating axis (Figs. S8C and D).
The complete motion is shown in Movie S3. 
As analyzed in Fig. S9, SMOL results perfectly match to the optical ground truth with an average error below 0.2 mm.
Propulsion directions (red arrows) at the respective locations deviate less than 5° from the optically determined direction.

A strong magnetic field, necessary for actuation, disrupts the SMOL method, as it saturates the magnetic sensors and also acts as additional damping torque to force the cantilever's oscillation to stop.
Therefore, the actuation and localization have to be performed alternatingly. 
Since the torque-based robot (Fig. 3B) utilizes a rotating permanent magnet for actuation, the sample container holding the robot and the gel needs to be transferred between the actuation and the localization setups.
For the gradient-based robot (Fig. 3A), the actuation field can be shut off quickly, resulting in high refresh rates for closed-loop control in a single setup.
These results show the accurate localization possibility with the SMOL method. Moreover, they also demonstrate the feasibility of using the same magnetic moment on the miniaturized robot for both actuation and localization purposes in various media.

\subsection*{Integration of SMOL for biomedical applications}

Accurate navigation of surgical tools in the human body is essential for the success of medical procedures.
Low-frequency magnetic fields ($<$100 kHz) are able to penetrate biological tissues with no noticeable attenuation due to the negligible difference of the magnetic permeability between air and water \cite{tenforde1991}.
Additionally, magnetic fields are well-known for their bio-compatibility even beyond a field-strength of 1 T for MRI machines \cite{schenck2000}.
In comparison, the magnetic field generated by the SMOL device's magnet is in the micro- to nanotesla range, and the external field for excitation is in the millitesla range.

For the demonstration of surgical tool localization and magnetic tool compatibility, a SMOL device is attached to the front of a flexible endoscope that is manually navigated through a transparent {\it{in vitro}} kidney organ phantom \cite{adams2017} as a commonly performed procedure in retrograde minimally-invasive intra-renal surgeries \cite{inoue2021}, shown in Fig. 4A and B.
The endoscope itself is weakly magnetic and produces a surface magnetic field of maximally 120 µT, which makes it compatible with the sensors at distances larger than 2 cm.
The endoscope tip first navigates towards the lower calyx, then towards the upper calyx and is finally retracted from the kidney.
Meanwhile, the tip of the tool is tracked with an average 4.2 Hz refresh rate using the attached SMOL device (Movie S4).
Since the orientation of the SMOL tracker changes during sharp bending of the endoscope tip (over 180°), the excitation field orientation is adapted according to the last known orientation of the device.
Localization of the endoscope within the whole kidney is achieved in real-time with full position and rotation information, without the use of radiation or optical methods.
A further demonstration of the high tolerance of SMOL to magnetic interference, \textit{e.g.} by a surgical tool, is shown in Fig. S10, where additionally a direct comparison with static permanent magnet localization is performed.

Besides the tracking of mobile millirobots and surgical tools, a SMOL device can also be used as a static biomedical marker.
Strong mechanical damping behavior is expected to impede mechanical oscillations in soft biological tissues, which creates an additional challenge to obtain a sustained oscillating signal of the SMOL device for an accurate localization.
Brain is known as one of the softest tissues in the human body \cite{budday2020}, therefore, this tissue is chosen as a realistic and strict testing environment for the SMOL method.
The results of the SMOL method working in an \textit{ex vivo} porcine brain are shown in Fig. 4C-E.
The SMOL device is implanted into the gray matter in the cerebrum (Fig. 4C) and successive US imaging (Fig. 4D) is used to obtain planar information of the location of the device relative to the rigid boundaries of the sample container.
As shown in the image, the overall US imaging resolution and contrast are poor due to multiple reflections and scattering of the US beam in inhomogeneous biological tissues.
The tracker at a distance of $\sim$40 mm to the US probe is barely distinguishable from the background noise.
On the contrary to US imaging, the SMOL method accurately detects the position and orientation of the robot in the brain.
When the localization information is overlaid with the US image (Fig. 4E), a very good correlation is found between the two localization methods and the tracker's major axis (red arrow) aligns very well with the estimated orientation by US imaging.
At 45 mm $z$-distance from the magnetic sensors, the standard deviation of 10 independent measurements is in the sub-millimeter region, which reveals the high usability of the SMOL method in real biological soft tissues.

\section*{Discussion} 

\subsection*{Advantages of the SMOL method} 

SMOL offers a completely wireless localization method of small-scale devices in magnetically noisy and highly-damped biological environments.
It outperforms state-of-the-art magnetic localization techniques in many aspects, and the use of a single magnetic moment and a cantilever for the restoring force brings unique benefits.

First, the small footprint of the SMOL device enables its integration into small-scale robots and minimally-invasive surgical tools, \textit{e.g.} endoscopes and possibly catheters and needles in the future.
It requires no on-board power, which makes it easier to be integrated with wireless medical devices, such as capsule endoscopes and implants.

Second, the unique frequency response of a SMOL device facilitates the isolation from DC and low frequency magnetic noise, \textit{i.e.} surrounding magnetic and electronic devices or moving surgical tools.
The resonance frequency is readily tunable by the material and the geometry of the cantilever.
Precise manufacturing techniques are required for frequency tuning with the environmental noise in consideration.

Third, the high accuracy for all 6 DoF over large distances in addition to its small size are beyond the possibilities of other wireless tracking methods.
Especially, most state-of-the-art methods based on static permanent magnets, allowing actuation and 5 DoF localization with the same magnet, do not achieve sub-millimeter accuracy at lower or equal localization depths, and are using multiple orders of magnitude larger magnetic volumes \cite{son2016,schlageter2001,guitron2017,xu2022,song2016,nicolae2019,xu2019}.
Owing to the physical principle behind the evaluation method, the accuracy is also highly tuneable by over one order of magnitude from 0.47 mm / 2.14° down to 40 µm / 0.15° (see Fig. 2D), allowing a localization accuracy of less than one-tenth of a millimeter at large depths for very demanding localization applications.

Fourth, SMOL is proven to work in various boundary conditions. Although physical boundaries have large impact on the attenuation of the mechanical oscillation, experiments show the principle works with solid boundaries, soft viscoelastic materials and even liquid boundaries.
The stronger coupling to the surrounding environment is also beneficial for material sensing applications \cite{fischer2022}.
Moreover, SMOL requires no physical contact of the external device to soft tissues, which will benefit minimally-invasive and robotic surgeries, where a direct contact of the imaging probe to the internal organs is often not possible.

\subsection*{Integration considerations for robotic applications}

The movement freedom required for the magnet's oscillation necessitates careful considerations with respect to the desired application.
The helical robot (Fig. 3B) is actuated by converting torque about the propulsion axis into linear translation by the screw-shaped housing.
The torque from an external field, however, is only applied to the magnet, which is attached to the delicate cantilever.
If the magnet is free to rotate without angular restriction, the strong magnetic torque will keep twisting the cantilever and exceed the strength limit of the cantilever material, leading to permanent plastic deformation or fracture of the beam.
Hence, geometric constrains are added inside the housing to transmit the torque by the direct contact to the housing.
Furthermore, depending on the angular speed of the external actuation B-field, mechanical instabilities, appearing in the form of path deviations or wobbling, in helical robots can occur \cite{mahoney2012}.
This behavior may be amplified in the SMOL helical robot, since the magnet freely deflects according to the external fields, leading to misalignment of the magnet with the robot's central axis.

Most non-magnetic localization methods cannot provide more than three spatial coordinates, which is very limiting for robotic control, whereas localization method based on single permanent magnets can maximally achieve 5 DoF.
As the magnetic field exhibits rotational symmetry around the axis of the magnetic moment, it is fundamentally impossible to measure the rotation angle of a robot around this axis.
Lack of information about this axis can lead to control difficulties, for example when using magnetic helical robots or wireless endoscopy capsules, since the propulsion axis is perpendicular to the magnetic moment vector and therefore unknown.
Other means of determining the sixth DoF require knowledge of previous locations \cite{xu2022}, which are vulnerable to interruptions or sudden turning, or additional embedded sensors \cite{popek2017,taddese2018}, which drastically increase the device size.
The integration of magneto-oscillatory methods such as SMOL, as shown in Fig. 3, circumvents the need of additional implementations by directly providing all 6 DoF.
Besides the need for 6 DoF localization for closed-loop actuation, it can also provide crucial information for the navigation of endoscopes.
The tip of an endoscope is equipped with a camera, which provides optical feedback for the surgeon, hence the navigation precision and interpretation of the endoscopic images can be significantly enhanced by full rotation information. 

Closed-loop control in real-time is desired for most robotic systems.
In Fig. 3A, the capabilities of SMOL to fulfill this demand are shown.
Current limitations of the localization rate lie in the saturation of the magnetic sensors during the excitation, meaning that excitation and sensing cannot be performed simultaneously.
A five cycle excitation for a SMOL device with $f_\mathrm{res} = 103.5$ Hz, in addition to the coil ringdown, requires around 80 ms.
For a single cycle excitation, it could be reduced to below 40 ms. 
Realistically, by reducing the coil ringdown and computation time, a localization refresh rate above 25 Hz can be expected for the SMOL method, making it very suitable for closed-loop robotic applications.
Exploiting the physical principle allows superfast localization rates even beyond 50 Hz, revealing the full potential of the SMOL method for real-time biomedical and robotic applications.

The magnet for the used SMOL devices occupies a volume of 0.8 mm³, and the device as a whole (Fig. 1 and Fig. S7) requires approximately 14 mm³ of space, owing to the volume needed for cantilever deflection.
The minimal footprint makes it ideal for a direct integration into existing applications such as endoscopic capsules, surgical tools or as a stand-alone minimally-invasive fiducial marker. 
The magnetic signal strength, and therefore the localization distance, linearly scales with the magnets size over the distance.
Hence, depending on the applications' needs, the device could be scaled to achieve larger distances or smaller sizes.
A shorter cantilever, for example, could allow a larger deflection angle and a stronger signal, and the resulting resonance frequency increase could be compensated by thinning of the beam.
In general, up-scaling requires a spread-out sensing array, while down-scaling poses additional challenges to manufacturing which could be overcome with MEMS fabrication methods.

The working volume of the SMOL method is dominated by the excitation coils.
Magnetic fields in all directions and at all positions within the desired volume have to be generated for excitation of an arbitrarily oriented SMOL device.
Currently, two perpendicular excitation coils are used (Fig. S1A), which can achieve excitation for all in-plane orientations, as demonstrated by the closed-loop controlled robot in Fig. 3A.
The excitation coil can be extended to an optimized 3-axis multi-coil for excitation of any orientation of the SMOL device.
For extension of the actuation system from 2D to 3D, which has been demonstrated for many small-scale magnetic robots in the past \cite{ebrahimi2020}, additional coils above and below the working plane are needed to overcome buoyancy and gravitational forces, and their interference with the sensing array has to be minimized.

In summary, the SMOL method can be readily integrated with minimal volume occupation into many existing biomedical and robotic applications to offer high spatial and angular 6 DoF localization as well as real-time temporal resolution.

\section*{Materials and Methods} 

\subsection*{SMOL system components}

A wireless SMOL device consists of a housing and a mechanically resonant structure with an attached magnet.
The housing was designed and 3D printed to allow a sufficiently large deflection angle for the magnet ($\diameter$ 1 mm $\times$ 1 mm, grade N52). The magnet is attached to the tip of a steel stripe (C1095 spring steel, 20 µm thick, 0.2 mm wide, 3-5 mm long) to form the resonant cantilever structure, which is fixed to the housing. Details on the SMOL device fabrication can be found in Supplementary Materials.

System control and data processing were performed in MATLAB (R2022b, The MathWorks, US) and Python (V3.9.12, Python Software Foundation, US) on a i9-12900K 3.20 GHz processor with 64 GB RAM and an NVIDIA GeForce RTX 3060 GPU. 
For analog control and analog signal conversion, a data acquisition board (PCIe-6363, NI, US) with an input range of ± 11 V, 16 bit resolution (0.33 mV resolution) and a sampling rate of 50 kS/s was used.
The magnetic fields were measured with a 10-channel fluxgate sensor unit (FL1-10-10-AUTO, Stefan Mayer Instruments, Germany) with the individual sensors arranged in a 2D plane with a laser-cut (Beambox Pro, FLUX, US) polymethylmethacrylate (PMMA) plate. 
The sensor unit was equipped with a manual offset compensation function and a sensitivity level of 1 V/µT, a range of ± 10 µT, a resolution of 0.1 nT and an inherent noise of 20 pT/$\sqrt{\mathrm{Hz}}$ at 1 Hz.
For the excitation of the SMOL device, two perpendicular customized electromagnetic coils (0.56 mm-diameter enameled copper wire, $\sim$150 turns on a 60 mm $\times$ 50 mm $\times$ 10 mm and $\sim$150 turns on a 70 mm $\times$ 70 mm $\times$ 15 mm 3D printed bobbin) were built to generate a magnetic field above the coil ($\sim$1 mT at a distance of 30 mm) with full planar field-of-view.
They were powered by power amplifiers (TSA 4000, the t.amp, Germany) with a current amplitude of up to 10 A.
In order to prevent overheating, a customized water cooling reservoir was 3D printed which was filled with double-distilled water.
For fast ringdown times and removal of current-leakage, solid state relays (D2425-10, Sensata-Crydom, US) were used.
Between the end of the excitation phase and the start of the evaluation phase, a buffer time of 20 ms was applied to avoid the interference of the magnetic signal by the coil ringdown.
Statistical analysis was performed in MATLAB.
All experiments were performed at room temperature ($\sim$22°C) in an unshielded environment without any magnetic or electric shielding.

\subsection*{Closed-loop control of the gradient-actuated millirobot}

For the gradient-actuated millirobot, a planar array of electromagnetic coils was used (see Fig. S8A).
Four commercial coils (LSIP-330, Monacor, Germany) were arranged in a square with 80 mm side length, resulting in an effective distance of 84 mm between opposing coils.
They were powered by a 4-channel DC power supply (NGP800 with NGP K107 module, Rohde \& Schwarz, US), and controlled with four solid state relays (TC-GSR-1-25DD, Tru Components, Germany) in addition to the data acquisition unit.
The tank was filled with glycerol ($\geq$99\% purity, Honeywell, US). The millirobot partially submerged due to the buoyancy.
An additional external coil (EM-6723A, Pasco, US) operated with 400 mV and $\sim$20 mA DC (AFG31022, Tektronix, US) was placed at 150 mm distance to roughly compensate the earth's magnetic field in the working volume to avoid undesired alignment of the SMOL device with the earth's magnetic field.
A simplified closed-loop control scheme is shown in Fig. S8B and further details on the control as well as the fabrication procedure of SMOL millirobots, the numerical simulations, the accuracy characterization and the biomedical application demonstrations are presented in Supplementary Materials.

\subsection*{Helical millirobot actuation}

For the actuation of the helical millirobot, an external rotating magnetic field with an average magnetic field strength of $\sim$100 mT was used.
A cubic NdFeB magnet (side length 50.8 mm, N40, Supermagnete, Germany) was mechanically fixed to a stepper motor (23HS30-2804S, Stepperonline, US) with the magnetic axis perpendicular to the rotational axis of the motor.
The assembly was placed on a rotational stage (GFV5G50, Orientalmotor, Japan) to steer the rotational axis of the magnet in a 2D plane (see Figs. S8C and D).
The propulsion of the millirobot was restricted to a planar motion in the gel in-between a PMMA plate and the bottom of the container with a gap of 6 mm and the propulsion path was pre-defined in the gel for improved steerability.
As a viscoelastic testing environment to mimic the mechanical properties of a porcine brain \cite{navarro2019}, hydrogel with 3 wt.-\% gelatin and 0.2 wt.-\% agarose was used.
Both components were stirred together in double-distilled water at 80°C for 30 min, filled into rectangular plastic containers (50 mm $\times$ 60 mm $\times$ 15 mm) and cooled to 22°C for at least 4 h before use.
The rotation frequency of 0.25 Hz and the direction of the rotating magnetic field were controlled by an Arduino board (Ardunio Uno Rev3 SMD, Arduino), and the direction of the rotational axis was steered manually.
The propulsion was paused 8 times for the localization, and in each pause, the sample box (with the robot embedded statically inside) was removed from the actuation setup and put in the localization setup at the same location.
A camera system (EOS RP with a RF35mm F1.8 lens at 25 fps, Canon, Japan) was used to image the propulsion from the top view.
Eight videos were linked and analyzed by a customized code (MATLAB) to recognize the robot center-point and orientation in each frame and draw the trajectory to overlay on the original videos.


\section*{Acknowledgments} 

The authors are grateful to Prof. P. Fischer (University of Heidelberg) for fruitful discussions and acknowledge laboratory support by K. Jardak and D. Li.
F. Fischer thanks the International Max Planck Research School for Intelligent Systems (IMPRS-IS) for the support.
M. Jeong and T. Qiu acknowledge the support by the Stuttgart Center for Simulation Science (SimTech).
All authors acknowledge the support from the Institute of Physical Chemistry at the University of Stuttgart.

\paragraph*{Funding}
This work was partially funded by the Vector Foundation (Cyber Valley research group), the MWK-BW (Az. 33-7542.2-9-47.10/42/2) and the European Union (ERC, VIBEBOT, 101041975).

\paragraph*{Author contributions}

F.F., C.G. and T.Q. conceptualized the SMOL method.
F.F. and C.G. developed the data evaluation procedure.
C.G. coded the Levenberg-Marquardt optimization.
F.F. designed the experimental SMOL system, developed the mathematical model, fabricated the SMOL devices, collected and analyzed the data, and wrote the manuscript.
M.J. designed the torque-based actuation setup.
T.Q. supervised the project.
All authors contributed to the edition of the manuscript.

\paragraph*{Competing interests}
F.F., C.G. and T.Q. have a pending patent on the SMOL method (PCT / EP2023 / 072144).

\paragraph*{Data and materials availability}

All data needed to evaluate the conclusions in the paper are present in the paper or the Supplementary Materials.

\newpage

\section*{Figures} 
 \begin{figure*}[hp]
    \centering
    \includegraphics[width=\textwidth]{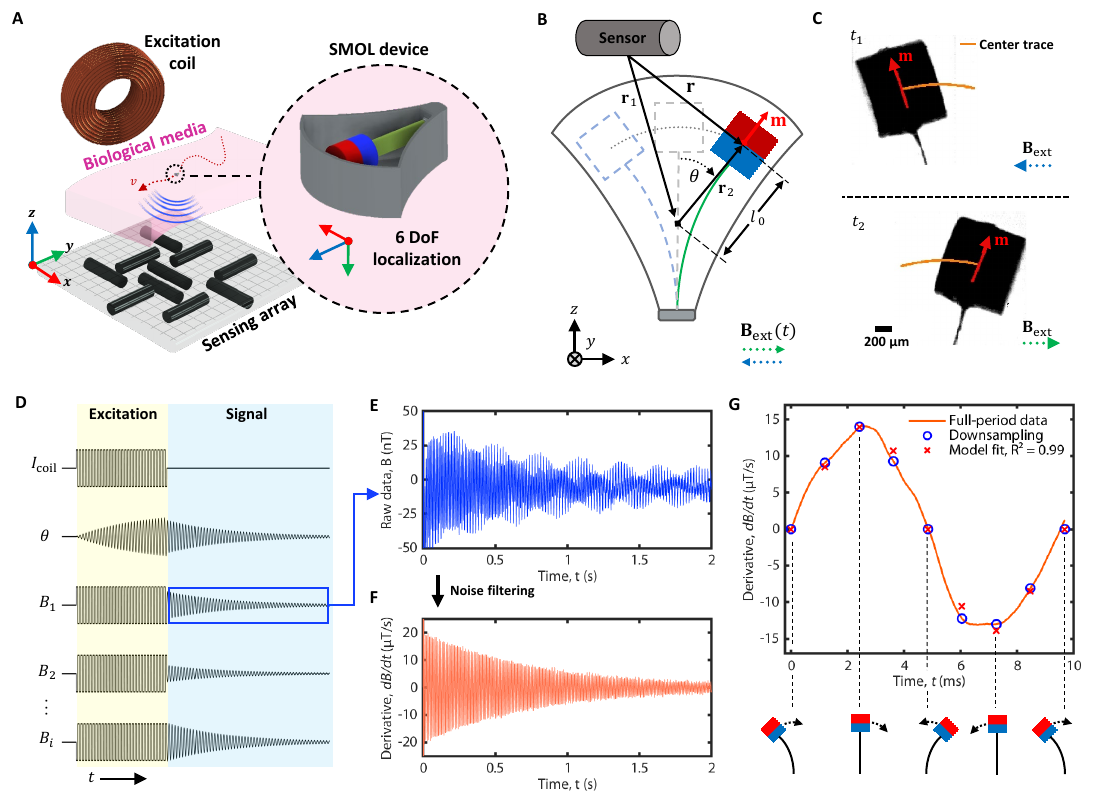}
    \caption{
    \textbf{Overview of the SMOL method.}
    (\textbf{A}) Schematic of a SMOL system with an excitation coil, an embedded SMOL device and a sensing array.
    (\textbf{B}) Oscillating cantilever model. The magnetic moment $\mathbf{m}$ at position $\mathbf{r}$ oscillates in time on a circular path upon excitation by an external B-field $\mathbf{B}_\mathrm{ext}$.
    The deflection angle $\uptheta$ is limited by the housing. 
    (\textbf{C}) High-speed optical analysis of the magnets center point (orange) and orientation (red) at two times during the oscillation.
    (\textbf{D}) Schematic illustration of the excitation and signal phase.
    During the excitation phase, a current $I_\mathrm{coil}$ is applied to the excitation coil.
    The resulting B-field leads to a continuous increase of the deflection angle $\uptheta$ and to the saturation of the sensor signals $B_1$ to $B_i$. After $I_\mathrm{coil}$ is shut down, $\uptheta$ slowly decays in an underdamped harmonic oscillation and the sensors measure the signal emitted from the SMOL device. 
    (\textbf{E}) Raw signal measured at 80 mm distance with a resonance frequency $f_\mathrm{res} = 103.5$ Hz.
    (\textbf{F}) Resulting signal after filtering.
    (\textbf{G}) Full oscillation period. The corresponding cantilever deflection is illustrated below.
    The fully sampled data (orange) is down-sampled to 5 points per half-period (blue) for insertion into the optimization model.
    The fitting result (red) for the physical model yields an excellent fit with R² = 0.99.
    }
    \label{fig:1}
  \end{figure*}
  

  \begin{figure*}
    \centering
    \includegraphics[width=\textwidth]{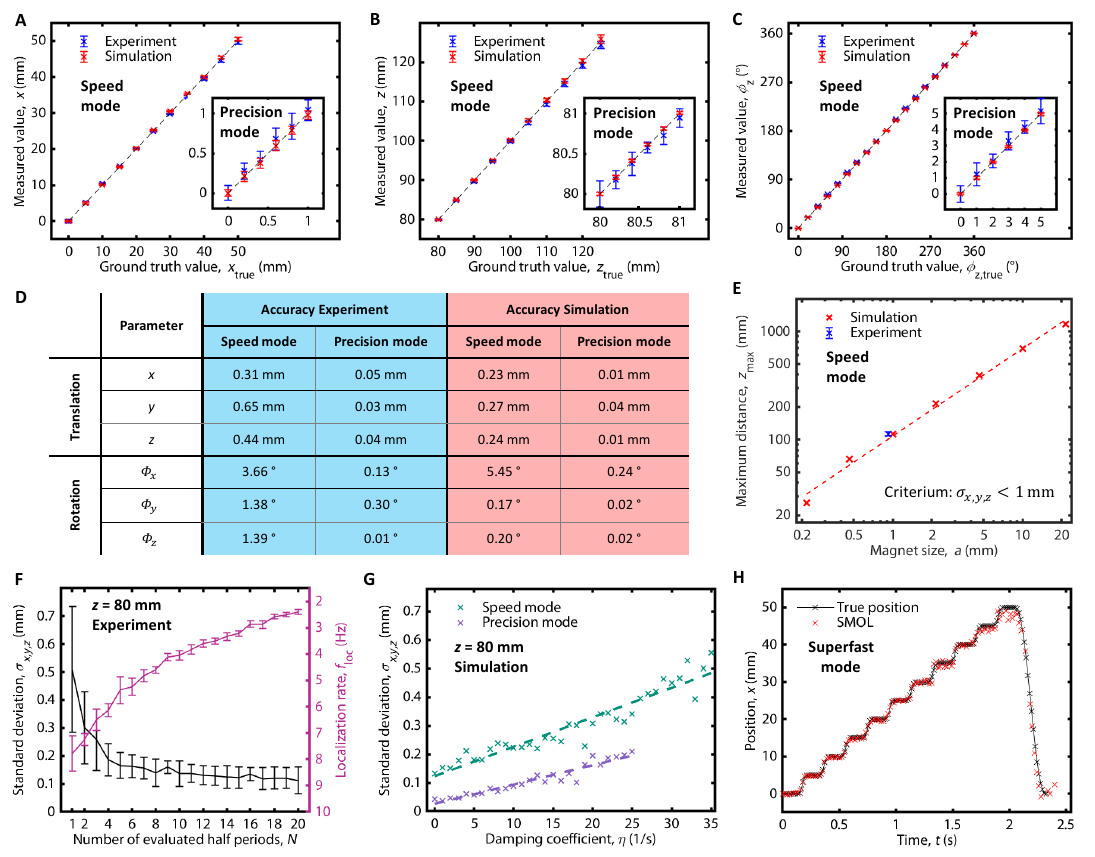}
    \caption{
    \textbf{Characterization of the 6 DoF accuracy and localization rate.}
    (\textbf{A}--\textbf{B}) Translational accuracy in $x$- and $z$-direction for the speed mode (number of evaluated half periods $N = 2$) and precision mode ($N = 20$) as inset, respectively.
    Experiments (blue) and simulations (red) accurately represent the ground truths, as the data points span on the 45° line in the graphs.
    (\textbf{C}) Rotational accuracy around the $z$-axis, which is the missing rotational DoF for a static magnet (see Fig. 1B).
    (\textbf{D}) Summary table for the accuracies in presented ranges.
    The translation axes correspond to the systems extrinsic axes with respect to the sensor array (Fig. 1A), while the rotation axes are the intrinsic axes of the SMOL device according to Fig. 1B.
    (\textbf{E}) Maximum localization distance $z_\mathrm{max}$ versus magnet size $a$ of an equivalent cube.
    (\textbf{F}) Standard deviation $\sigma_{x,y,z}$ and localization rate $f_\mathrm{loc}$ versus $N$.
    (\textbf{G}) $\sigma_{x,y,z}$ versus damping coefficient $\eta$ (Eq. S4) for both modes with linear trend lines. A larger $\eta$ indicates a faster decay of the signal.
    (\textbf{H}) {\it Superfast} localization demonstration with 51.7 Hz refresh rate for a linear stepping motion. The original signal is segmented into $N_\mathrm{seg} = 4$ half period segments.
    }
    \label{fig:2}
  \end{figure*}


  \begin{figure*}
    \centering
    \includegraphics[width=\textwidth]{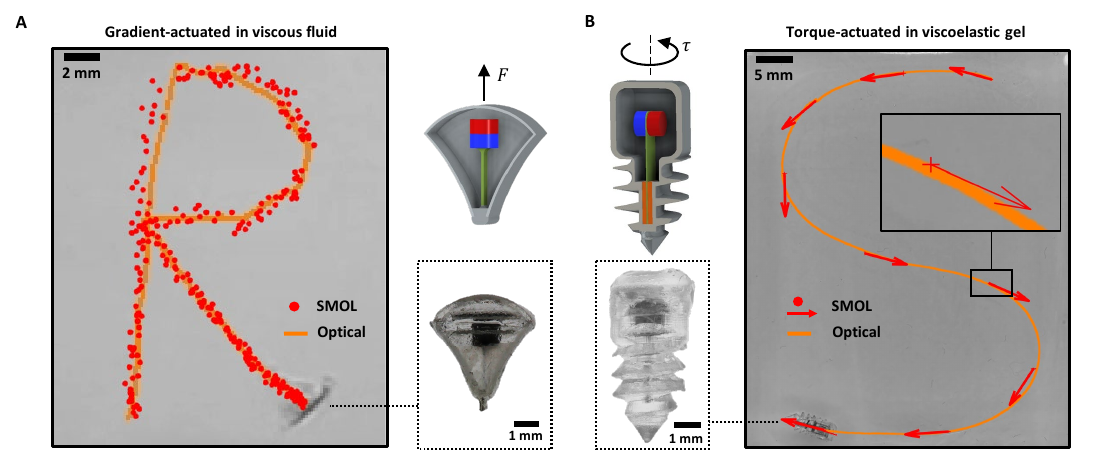}
    \caption{
    \textbf{Integration of SMOL and actuation in millirobots.}
    (\textbf{A}) \textit{R}-shaped actuation path of a millirobot in viscous fluid determined by optical tracking (orange) and SMOL tracking (red).
    The robot is controlled in a closed-loop by a magnetic gradient setup (Figs. S8A and B) and localized using SMOL at an average refresh rate of 3.5 Hz.
    Note that optical tracking was not used as feedback and is only presented for reference.
    (\textbf{B}) \textit{S}-shaped actuation path of a helical millirobot in viscoelastic gel. Red arrows indicate the main axis of the robot determined by SMOL.
    An external rotating magnet (see Figs. S8C and D) induces a torque $\tau$ along the main axis of the robot due to the perpendicular alignment between the cantilever (green) and the magnet (blue-red).
    }
    \label{fig:3}
  \end{figure*}
  
  \begin{figure*}
    \centering
    \includegraphics[width=\textwidth]{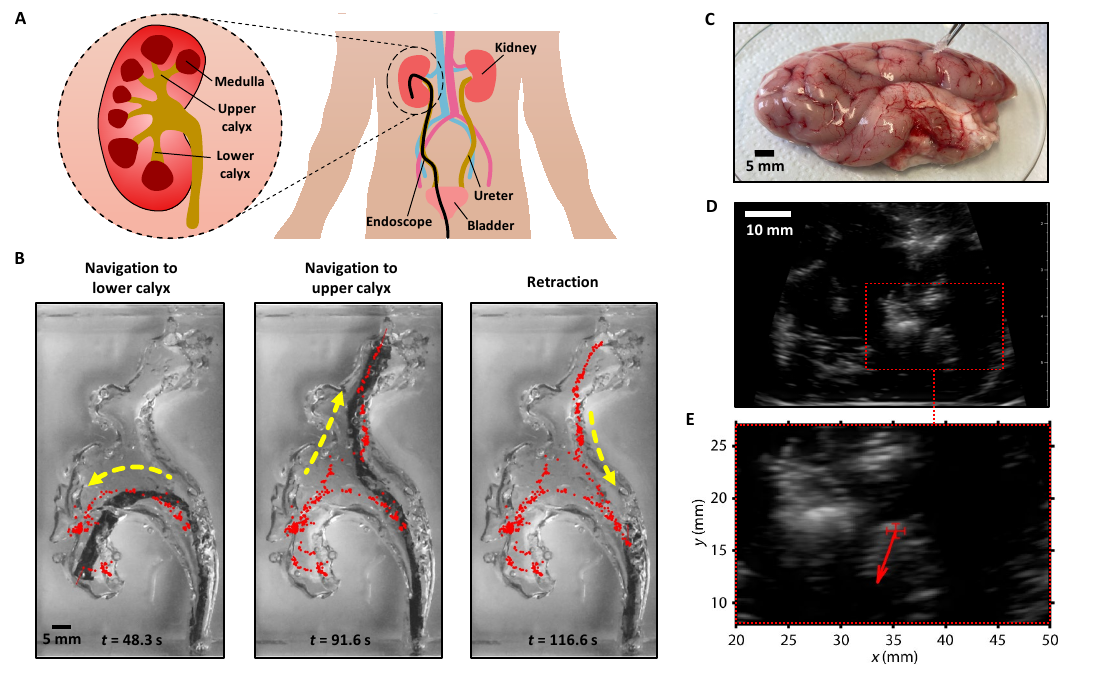}
    \caption{
    \textbf{Integration of SMOL for biomedical applications.}
    (\textbf{A}) Schematic illustration of the urinary tract.
    (\textbf{B}) Navigation of a flexible endoscope with an attached SMOL device in an {\it{in vitro}} kidney organ phantom.
    The localization results of the endoscopic tip by SMOL are shown as red dots, and yellow arrows indicate the movement of the endoscope.
    An average localization rate of 4.2 Hz was achieved (see Movie S4).
    (\textbf{C}) Implantation of a SMOL device into an {\it{ex vivo}} pig brain.
    (\textbf{D}) Ultrasound (US) image of the implanted SMOL device.
    A distinction of the tracker to the background is difficult due to biological inhomogeneity.
    (\textbf{E}) The overlay of the SMOL result on the enlarged US image.
    The position and orientation (red) are precisely determined by SMOL.
    }
    \label{fig:4}
  \end{figure*}
  


\newpage
\maketitle  

\section*{Supplementary Materials} 

\setcounter{equation}{0}
\setcounter{figure}{0}

\baselineskip16pt 
\begin{itemize}
  \item Movie S1: SMOL system overview and high-speed recording of a SMOL device with a resonance frequency of 135 Hz (5000 fps recording, 41$\times$ slowed) with 10 excitation cycles.
  \item Movie S2: Recording of a gradient-pulled SMOL millirobot being steered in a viscous fluid.
  The visual path (orange) and the SMOL positions (red) are overlaid.
  The movie is sped up for ten times.
  \item Movie S3: Recording of a torque-actuated helical SMOL millirobot being propelled and steered in a hydrogel. The visual path (orange) and the SMOL positions and orientations (red) are overlaid.
  The movie is sped up for twenty times.
  \item Movie S4: Recording of a flexible endoscope with an attached SMOL device being manually controlled inside an {\it{in vitro}} kidney organ phantom.
The SMOL positions (red-purple) are overlaid.
The movie is sped up for five times.
  \item Fig. S1: SMOL system components.
  \item Fig. S2: High speed optical analysis of a SMOL oscillation.
  \item Fig. S3: Signal filtering process.
  \item Fig. S4: Accuracy for the three translational DoF.
  \item Fig. S5: Accuracy for the three rotational DoF.
  \item Fig. S6: Superfast localization results.
  \item Fig. S7: Housing and lid of the helical and gradient-pulled millirobot.
  \item Fig. S8: Magnetic actuation setups.
  \item Fig. S9: Path analysis of millirobot demonstrations.
  \item Fig. S10: Tolerance of SMOL to magnetic interference by a surgical tool, in comparison to static magnet localization.
  \item Supplementary Materials and methods.
\end{itemize}
\baselineskip24pt 

\newpage

\subsection*{Supplementary figures}
\renewcommand{\thefigure}{S\arabic{figure}}

\begin{figure*}[hp!]
    \centering
    \includegraphics[width=1\textwidth]{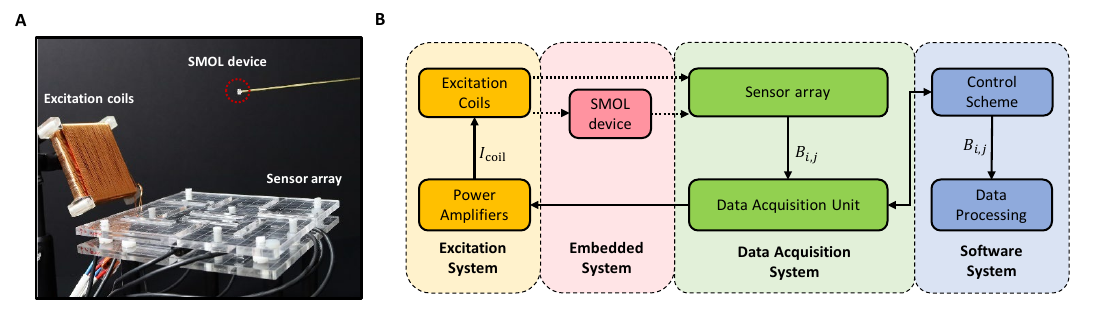}
    \caption{
    \textbf{SMOL system components.}
    (\textbf{A}) Main components of the SMOL system. The SMOL device (circled in red) is attached to a rod for visibility. 
    (\textbf{B}) Block-diagram of the SMOL system with its four subsystems for software control and data analysis, data acquisition, excitation and the embedded system, which is the SMOL tracker. 
    }
    \label{fig:supp_real_system}
\end{figure*}

\begin{figure*}[hp!]
    \centering
    \includegraphics[width=1\textwidth]{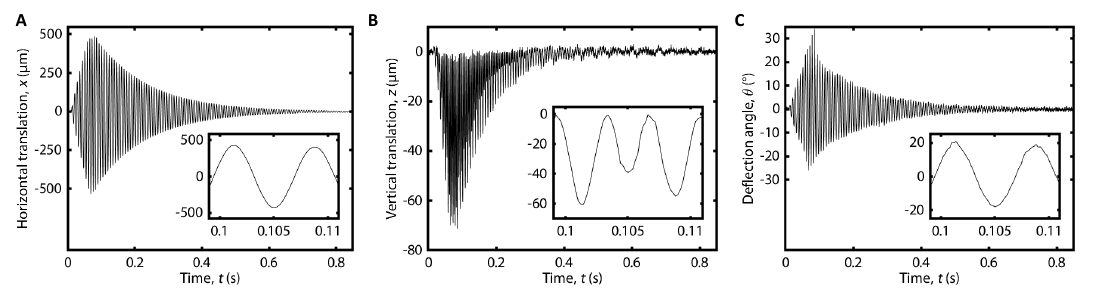}
    \caption{
    \textbf{High speed optical analysis of the SMOL oscillation.}
    Analysis of the optical center a SMOL device magnet with a resonance frequency of 135 Hz with its axes aligned as in Fig. 1B and 1C.
    The corresponding video is the supplementary Movie S1.
    (\textbf{A}) Horizontal translation $x$ over time $t$ with an inset showing a full period of the oscillation.
    (\textbf{B}) Vertical translation $y$ over time $t$ with an inset showing a full period of the oscillation.
    A double-frequency component occurs and its asymmetry can be attributed to optical misalignment.
    (\textbf{C}) Deflection angle $\uptheta$ of the magnets optical center over time $t$ with an inset showing a full period of the oscillation.
    Angles of over 20° are achieved for this specific SMOL device.
    }
    \label{fig:supp_high_speed}
\end{figure*}

\begin{figure*}[htbp]
    \centering
    \includegraphics[width=1\textwidth]{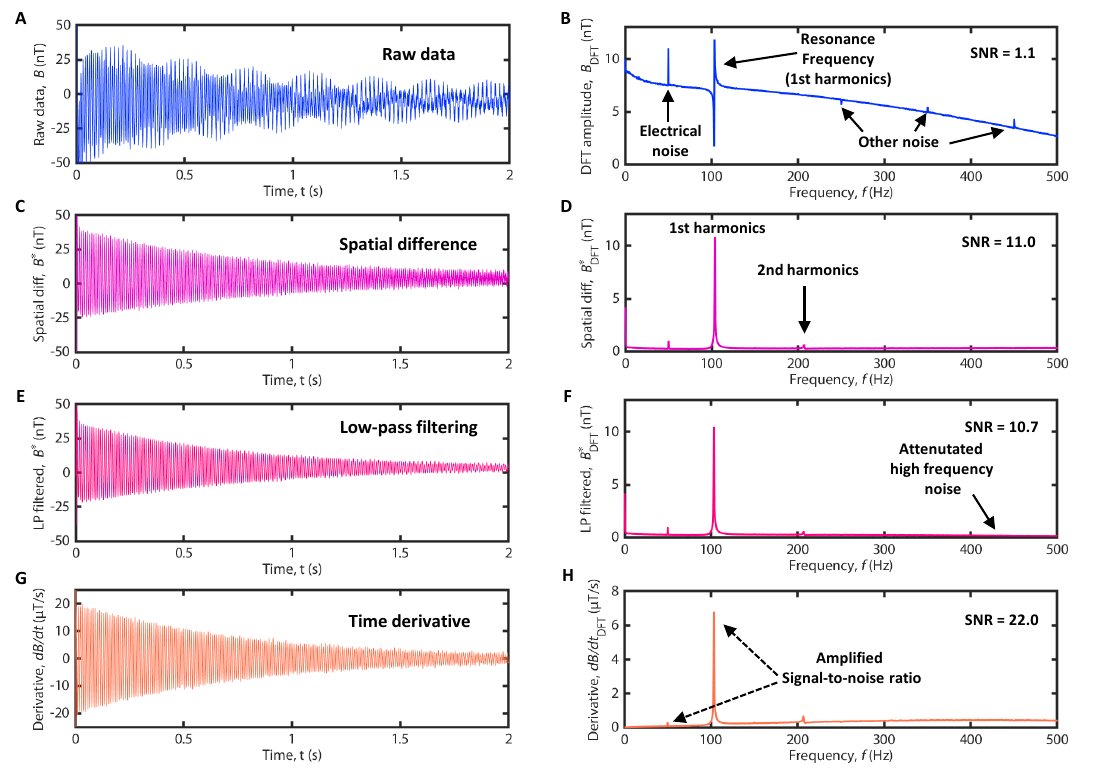}
    \caption{
    \textbf{Signal filtering process.}
    Signal filtering process for a SMOL signal recorded at 80 mm distance presented in the time-domain and frequency-domain. The SMOL device has a resonance frequency of 103.5 Hz. 
    (\textbf{A} - \textbf{B}) Raw signal and discrete Fourier transform (DFT) as recorded for 2 seconds (after the excitation phase) with a sample rate of 50000. The signal-to-noise ratio (SNR) across the whole signal with respect to 50 Hz noise amounts to 1.1. The SNR is approximated as the ratio between the 50 Hz amplitude and the resonance frequency amplitude.
    (\textbf{C} - \textbf{D}) Signal and DFT after subtracting the raw signal from the reference sensor of the same measurement direction. The remaining noise can be attributed to the spatial gradient of the noise. The SNR is increased ten-fold during this step to 11.0.
    (\textbf{E} - \textbf{F}) Signal and DFT after applying two successive moving mean filters with a window size of 50 (corresponding to 1000 Hz). This step removes high frequency jittering.
    (\textbf{G} - \textbf{H}) Signal and DFT after calculating the central time difference. Low frequencies will be attenuated while high frequencies will be amplified. The final SNR obtained by the shown filtering methods is 22.0. 
    }
    \label{fig:supp_real_FFT}
\end{figure*}

\begin{figure*}[htbp]
    \centering
    \includegraphics[width=1\textwidth]{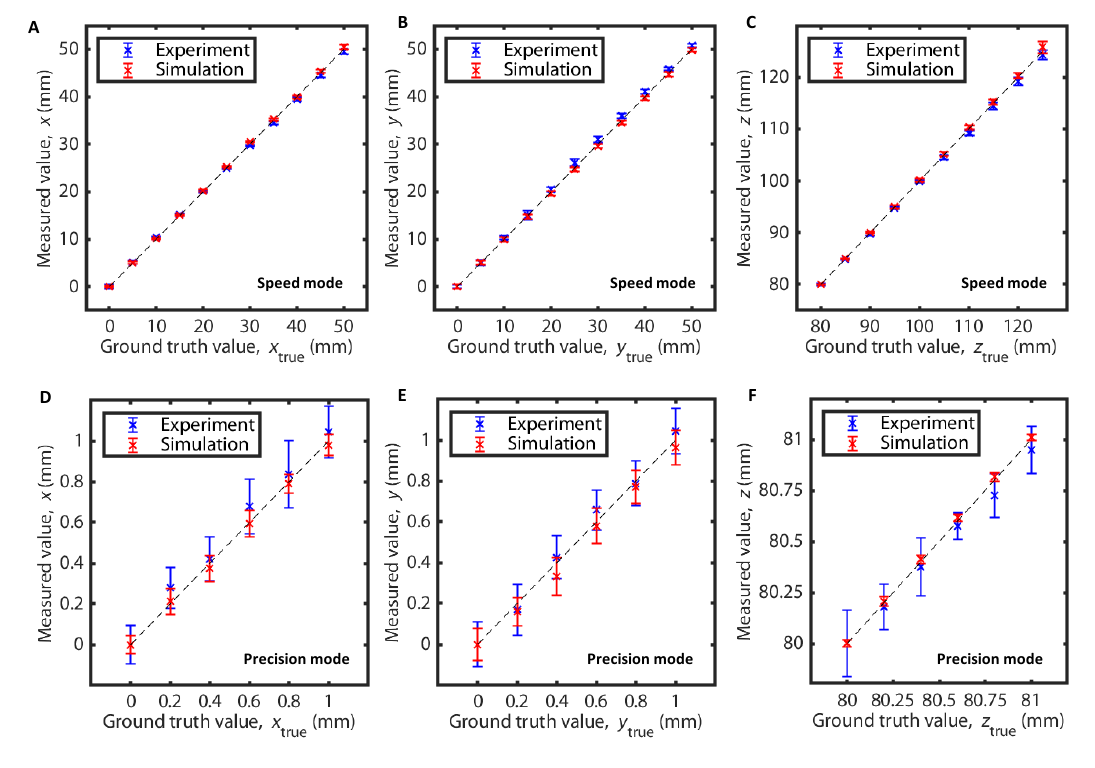}
    \caption{
    \textbf{Accuracy for the three translational DoF.}
    (\textbf{A} - \textbf{C}) Accuracy measurements for 50 mm translations in 5 mm increments along the $x$-, $y$- and $z$-axis in speed mode ($N = 2$, two half periods per signal are evaluated).
    (\textbf{D} - \textbf{F}) Accuracy measurements for 1 mm translations in 200 µm increments along the $x$-, $y$- and $z$-axis in precision mode ($N = 20$).
    Translation axes correspond to the systems extrinsic axes with respect to the sensor array. $x$ and $y$ translations are planar movements and $z$ corresponds to the distance to the sensor plane.
    }
    \label{fig:supp_accuracy_trans}
\end{figure*}

\begin{figure*}[htbp]
    \centering
    \includegraphics[width=1\textwidth]{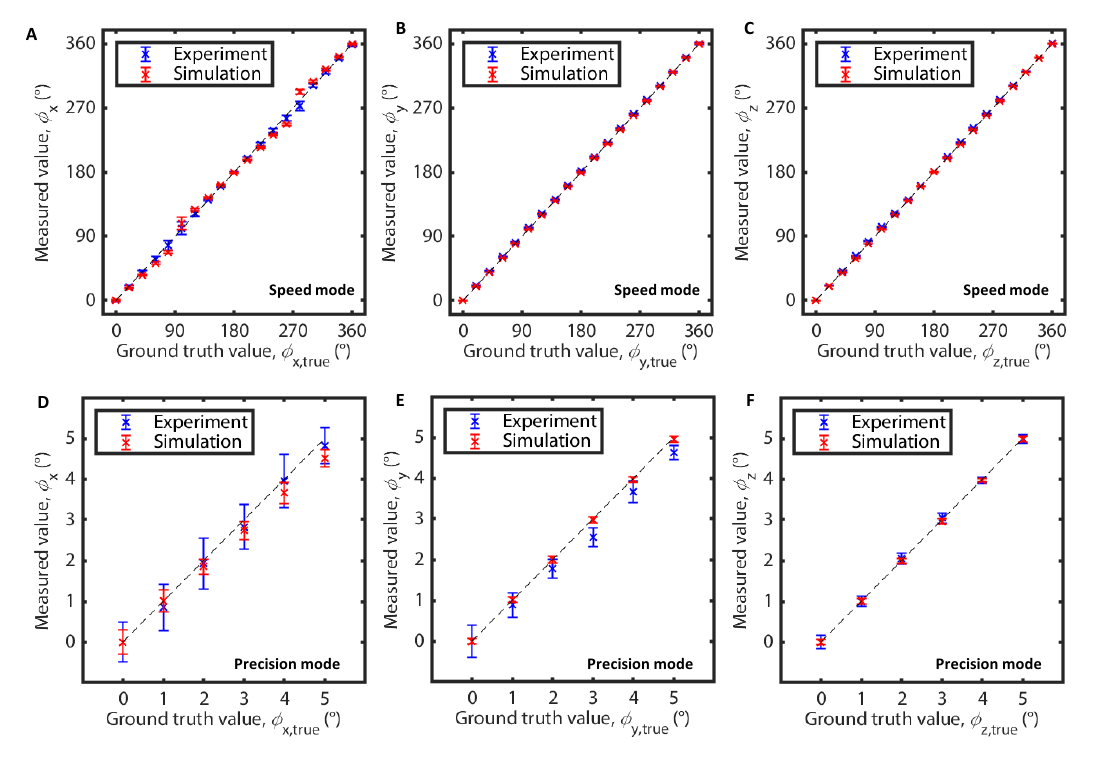}
    \caption{
    \textbf{Accuracy for the three rotational DoF.}
    (\textbf{A} - \textbf{C}) Accuracy measurements for full 360° rotations in 20° increments around the $\phi_x$-, $\phi_y$- and $\phi_z$-axis in speed mode ($N = 2$, two half periods per signal are evaluated).
    (\textbf{D} - \textbf{F}) Accuracy measurements for 5° rotations in 1° increments around the $\phi_x$-, $\phi_y$- and $\phi_z$-axis in precision mode ($N = 20$).
    The presented rotation axes are the intrinsic axes of the SMOL device corresponding to Fig. 1B. 
    }
    \label{fig:supp_accuracy_rot}
\end{figure*}

\begin{figure*}[htbp]
    \centering
    \includegraphics[width=1\textwidth]{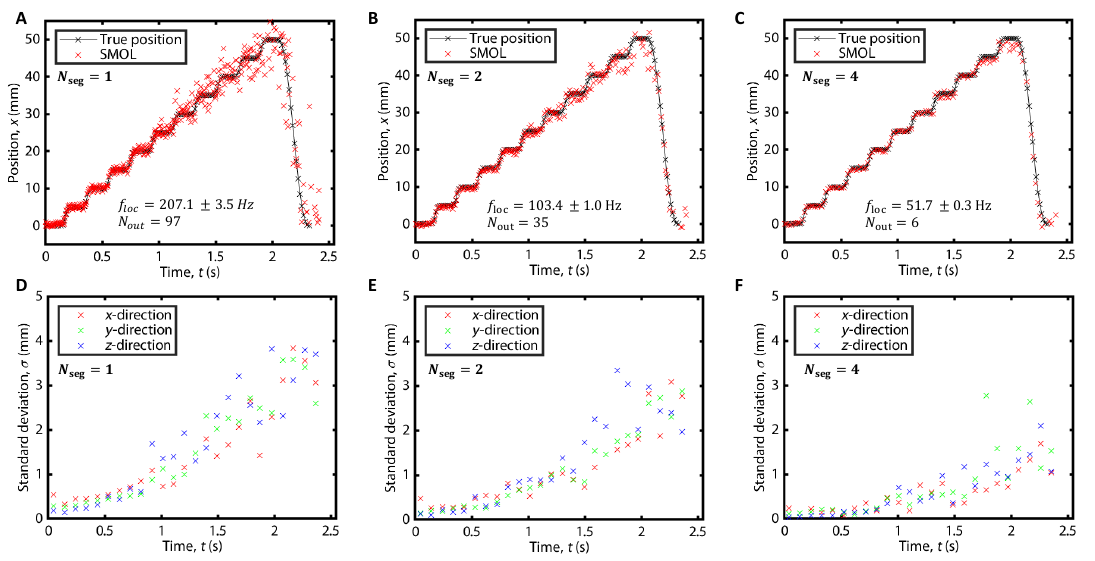}
    \caption{
    \textbf{Superfast localization results.}
    {\it{Superfast}} localization for a 50 mm linear motion with 5 mm stepping increments. The SMOL device resonance frequency is 103.5 Hz and the signal stemming from only one excitation is used. All values are calculated after complete signal acquisition.
    (\textbf{A} - \textbf{C}) True positions of the translation stage (black) and SMOL results (red) over time $t$ for different numbers of half periods per segment $N_\mathrm{seg}$.
    The achieved localization rates $f_\mathrm{loc}$ and number of outliers $N_\mathrm{out}$ are indicated in each graph.
    Outliers are defined by $x<$ -1 mm and $x>$ 60 mm and occur mainly in the last third of the shown time range.
    (\textbf{D} - \textbf{E}) Standard deviation $\sigma$ in the three spatial directions $x,y,z$ over time $t$ for a static location of the SMOL device for differing $N_\mathrm{seg}$.
    Generally, the later the evaluation segment, the lower the precision due to a decreasing signal-to-noise ratio.
    For larger $N_\mathrm{seg}$, the precision is effectively improved by noise averaging at the cost of localization rate.
    A bin size of $20/N_\mathrm{seg}$ for the calculation of $\sigma$ is used.
    }
    \label{fig:supp_superfast}
\end{figure*}

\begin{figure*}[htbp]
    \centering
    \includegraphics[width=1\textwidth]{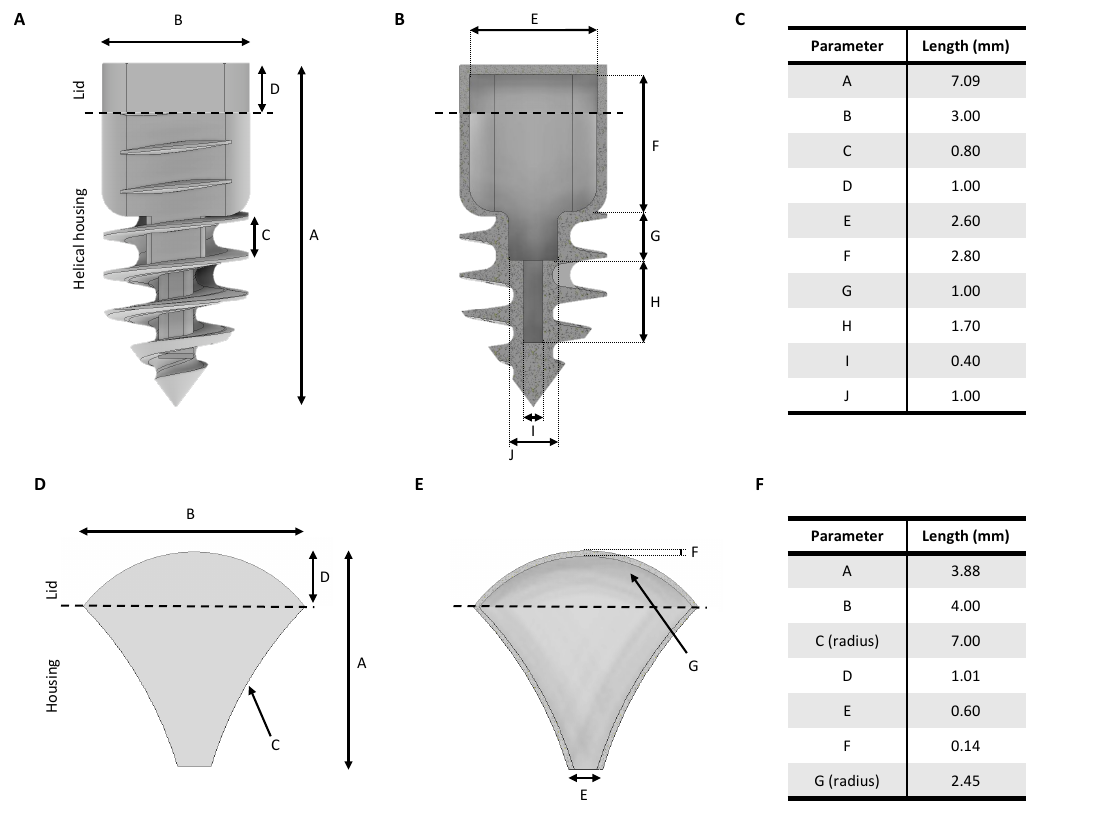}
    \caption{
    \textbf{Housing and lid of the helical and gradient-pulled millirobot.}
    (\textbf{A-C}) Side view, cross-section view and parameter table, respectively, of the helical-shaped driller millirobot used for actuation in a viscoelastic gel.
    (\textbf{D-F}) Side view, cross-section view and parameter table, respectively, of the gradient-pulled millirobot used for actuation in a viscous fluid.
    }
    \label{fig:supp_CAD}
\end{figure*}

\newpage

\begin{figure*}[htbp]
    \centering
    \includegraphics[width=1\textwidth]{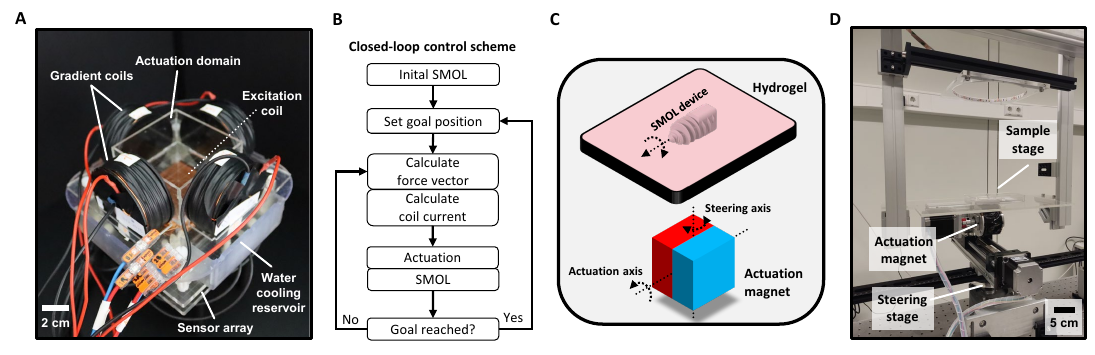}
    \caption{
    \textbf{Magnetic actuation setups.}
    (\textbf{A}) Experimental setup for closed-loop actuation of the gradient-pulled millirobot comprising four gradient coils for actuation. The excitation coil and sensor array are used for localization of the SMOL robot.
    (\textbf{B}) Scheme for the closed-loop control using SMOL as position and orientation feedback.
    (\textbf{C}) Schematic 2D actuation setup for the torque-based SMOL millirobot comprising an external permanent magnet with two rotation axes, one for steering and one for actuation. The robots rotation in the opposite direction as the actuation magnet.
    (\textbf{D}) Experimental setup for 2D actuation and steering of the helical millirobot.
    }
\label{fig:supp_actuation_system}
\end{figure*}

\begin{figure*}[htbp]
    \centering
    \includegraphics[width=1\textwidth]{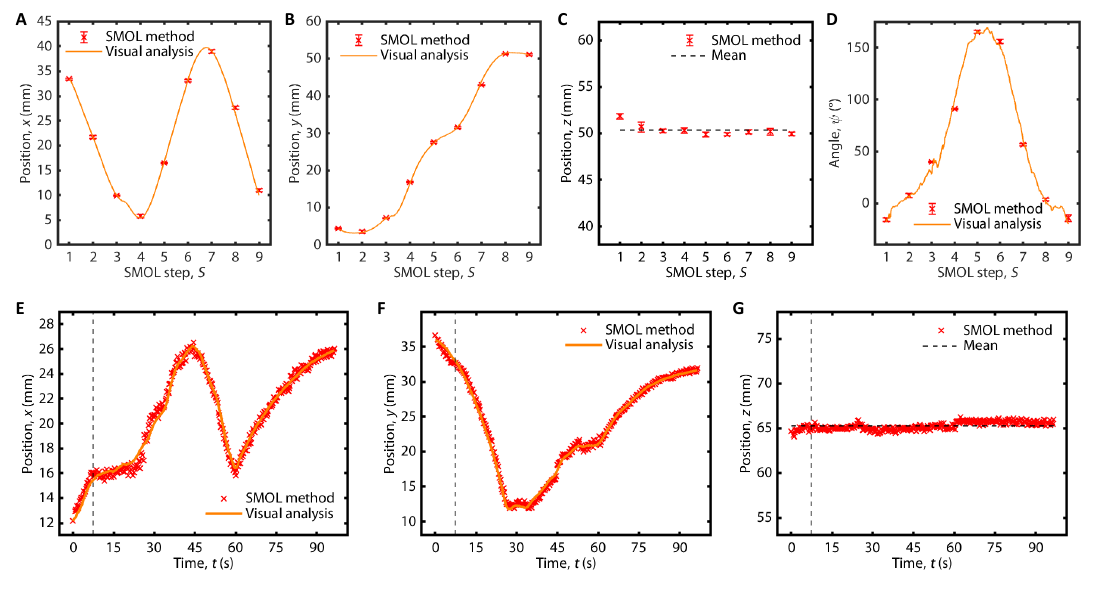}
    \caption{
    \textbf{Path analysis of millirobot demonstrations.}
    (\textbf{A-D}) Spatial positions $x$,$y$,$z$ and the propulsion axis angle $\psi$ of the torque-actuated helical millirobot, moving in an {\it{S}}-shape, compared between SMOL at steps $S$ and optical tracking.
    The mean position error to the optical path ($x$ and $y$ position) amounts to amounts to 0.18 mm ± 0.15 mm and the mean angular error amounts to 4.7° ± 3.6°.
    The $z$-error from the mean value is 0.41 mm ± 0.44 mm, indicating a high depth stability.
    (\textbf{E-G}) Spatial positions $x$,$y$ and $z$ of the gradient-pulled millirobot, moving in an {\it{R}}-shape, compared between SMOL  and optical tracking at time $t$.
    A vertical line indicates the starting point of the pre-determined path. 
    The mean position error to the optical path ($x$ and $y$ position) amounts to amounts to 0.27 mm  ± 0.23 mm.
    The $z$-error from the mean value is 0.36 mm ± 0.22 mm.
    In total, 336 localization and actuation cycles were performed in 96.6 seconds, resulting in an average closed-loop control rate of 3.5 Hz. 
    }
    \label{fig:supp_millirobot_path_analysis}
\end{figure*}

\begin{figure*}[htbp]
    \centering
    \includegraphics[width=1\textwidth]{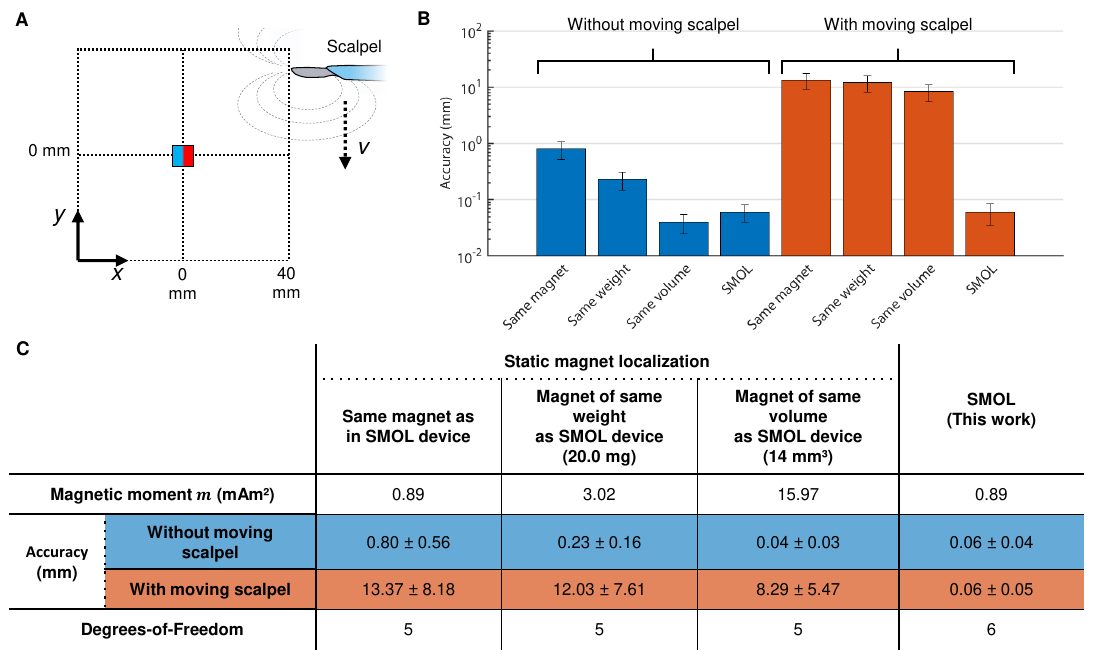}
    \caption{
    \textbf{Tolerance of SMOL to magnetic interference by a surgical tool, in comparison to static magnet localization.}
    (\textbf{A}) Schematic design of the simulation.
    A surgical scalpel (Swann Morton 10A), approximated as magnetic dipole with 15.97 mAm², is moving with 30 mm/s in $-y$ direction at $x = 40$ mm and $z = 80$ mm.
    The tracked device is located at the sensor array center with $z$-values of 80 mm to 100 mm in 5 mm increments.
    20 repetitions where performed per $z$-increment and at each repetition, the starting position of the moving scalpel was shifted by 4 mm in $-y$ direction.
   (\textbf{B}) Accuracy comparison for SMOL and various static permanent magnets with and without a moving scalpel. The SMOL method remains unaffected by the surgical tool while the static magnet accuracies drop by orders of magnitude.
   (\textbf{C}) Accuracy table for the data shown in Fig. S10B with the corresponding magnetic moment and degrees-of-freedom.
   Accuracies are evaluated as in Fig. 2D for the translation along the $z$-axis.
   Here, the signal duration per localization repetition is 100 ms for the permanent magnets and $N = 10$ for the SMOL device with same properties as for simulations in Fig. 2D.
    }
    \label{fig:supp_comparison_static_magnet}
\end{figure*}

\newpage

\subsection*{Supplementary Materials and methods}
\renewcommand{\theequation}{S\arabic{equation}}

\subsubsection*{Physical model for SMOL}
An external magnetic field $\mathbf{B}_\mathrm{ext}$ perpendicular to the magnetic moment of the magnet is applied to excite the oscillation. We assume solely a torque $\mbox{\boldmath$\tau$}$ is applied on the magnet that forces $\mathbf{m} $ to align towards the direction of $\mathbf{B}_\mathrm{ext}$: 
\begin{equation}\label{eq:magnetic_torque}
	\mbox{\boldmath$\tau$} = \mathbf{m} \times \mathbf{B}_\mathrm{ext}.
 \end{equation}
The torque is transmitted to the elastic beam that leads to a restoring torque (bending moment) and an angular deflection $\uptheta$. 
Due to physical boundaries of the available oscillation cavity (Fig. 1B), the cantilever beam with the magnet is constrained to a maximum angle of $\uptheta_\mathrm{max}$.
Using an alternating current (AC) magnetic field at the resonance frequency $f_\mathrm{res}$ of the mechanical structure as excitation, a periodic deflection can be induced, which decays as an underdamped harmonic oscillation in the form of:
\begin{equation}\label{eq:angle}
	\uptheta(t) =\uptheta_\mathrm{max}  \cos\left(2\uppi  f_\mathrm{res}  t + \upvarphi\right)\cdot D(t)
 \end{equation}
 with $\upvarphi$ as the phase and $D(t)$ as the time dependent damping term  ($0  \leq D(t) \leq 1$).
The latter is a monotonically decreasing function which can be approximated as linear
 \begin{equation}\label{eq:damping_lin}
     D(t) = -\upeta t + 1
 \end{equation}
 or exponential
 \begin{equation}\label{eq:damping_exp}
     D(t) = \exp(-\upeta t)
 \end{equation}
 with $\upeta$ as the damping coefficient.
 The linear approximation is found optimal for signal evaluations directly after the excitation with low damping while the exponential approximation is more suitable for rapidly damped signals in soft environments or signal evaluations later after the excitation.

The magnetic field stemming from a magnetic moment $\mathbf{m}$ is approximated by the ideal dipole model in Cartesian coordinates ({\it{51}}):
\begin{equation}\label{eq:dipole_cartesian}
	\mathbf{B}_\mathrm{cart}= \frac{\mu_0}{4\uppi \|\mathbf{r}\|^3} \left( 3 \hat{\mathbf{r}} \hat{\mathbf{r}} ^\top - \mathbf{I}\right)\mathbf{m},
 \end{equation}
where  $\mathbf{B}_\mathrm{cart} = \langle B_x,B_y,B_z\rangle $ are the magnetic field components at the position $\mathbf{r}$ with respect to the dipole center and $\mathbf{m}$ is the magnetic moment vector.
$\hat{\mathbf{r}}$ is the normalized vector of $\mathbf{r}$, $\mathbf{I}$ is the identity matrix and $\mu_0 $ is the permeability of free space. 
In relation to a point in space, here a magnetic sensor, the distance to the dipole center of an arbitrarily-oriented cantilever can be described by 
\begin{equation}\label{eq:r}
	\mathbf{r}= \mathbf{r}_1 + \mathbf{r}_2,
\end{equation}
where $\mathbf{r}_1$ is the vector between the sensor and the cantilever rotation center and $\mathbf{r}_2$ is the vector between the rotation center and the magnet, as presented in Fig. 1B.
$\mathbf{r}_1$ is constant for a fixed device position, whereas $\mathbf{r}_2$ is dependent on the current position of the magnet at time $t$.
While the magnet moves with the cantilever, a circular path of the magnet with a radius $l_0$ can be assumed and 2D polar coordinates in the $xz$-plane lead to
\begin{equation}\label{eq:r_CM}
	\mathbf{r}_2=  l_0 \mathbf{R}_q \langle\sin\left(\uptheta\left(t\right)\right),0,\cos\left(\uptheta\left(t\right)\right)\rangle
\end{equation}
with $l_0 = |\mathbf{r}_\mathrm2|$ being the length between the magnet center and the rotation center, $\mathbf{R}_q$ being a 3D rotation matrix (see Eq. S18) and $\uptheta$ being the deflection angle.
The time-dependent Eq. S5 is completed with the magnetic moment at rest pointing in positive $x$-direction:
\begin{equation}\label{eq:magnetic_moment}
	\mathbf{m}(t)= \mathbf{R}_q \mathbf{R}_y \frac{B_\mathrm{r} V}{\mu_0} \langle 1,0,0 \rangle.
\end{equation}
where $\mathbf{R}_y$ (see Eq. S19) describes the rotation around the $y$-axis for the angle $\uptheta$ as in a right-handed coordinate system (Fig. 1B), $B_r$ is the remanence field of the magnet and $V$ is the magnetic volume.
The time-dependent oscillation of the dipole moment is breaking the rotational symmetry of a static dipole around its magnetic moment axis by the rotation around a perpendicular axis (Eqs. S7 and S8), leading to a unique solution of Eq. S5 for the respective position $\mathbf{r}$ and rotation matrix $\mathbf{R}_q$.
Hence, all 6 DoF of the SMOL tracker can be determined.

Polar coordinates can be used to obtain a better understanding of the resulting magnetic signal from Eqs. S2 to S8. 
With radial component $\mathbf{r}$ and angular component $\uptheta$ (in radians), the magnetic field is defined by
\begin{equation}\label{eq:supp_Bpol}
     \mathbf{B}_\mathrm{pol} = \frac{||\mathbf{m}||}{||\mathbf{r}||^3}\langle 2\cos(\uptheta),\sin(\uptheta),0\rangle.
\end{equation}
The magnetic signal is a linear combination of the vector components in Eq. S9 and the deflection angle from Eq. S2 can be inserted, resulting in the full equation:
\begin{equation}\label{eq:supp_Bpol_full}
    B = \frac{||\mathbf{m}||}{||\mathbf{r}||^3}\left(2a\cos(\uptheta_\mathrm{max}  \cos\left(2\uppi  f_\mathrm{res}  t + \upvarphi\right)\cdot D(t)) 
    + b\sin(\uptheta_\mathrm{max}  \cos\left(2\uppi  f_\mathrm{res}  t + \upvarphi\right)\cdot D(t)) \right)
\end{equation}
with $a$, $b$ and $c$ as the respective scalar coefficients and $\sqrt{a^2+b^2+c^2}=1$.
Additionally, as shown in Eq. S6 to S7, $\mathbf{r}$ is also a function of $\uptheta(t)$.
Nested trigonometric functions are obtained which reveal the high complexity of the recorded signals.

\subsubsection*{Signal filtering}

As the SMOL method is utilizing complex magnetic signals to obtain position and orientation information from the device, careful filtering is necessary to not distort the true signal shape, shown in Fig. S3.
Noise can be removed most effectively by subtracting signals of sensors with the same orientation.
What remains is the desired signal's spatial difference $\mathbf{B}^\ast_\mathrm{S}$ in addition to the noise spatial difference $\mathbf{B}^\ast_\mathrm{N}$.
If $\mathbf{B}^\ast_\mathrm{S} >> \mathbf{B}^\ast_\mathrm{N}$, this filtering method is very efficient and, in the presented case, it increases the signal-to-noise ratio (SNR, with respect to 50 Hz noise) tenfold.
Sensor inherent electrical noise in the kHz range is smoothed out by moving mean filters.
Furthermore, as it is crucial to not distort the asymmetric signal shape, remaining low frequency noise is attenuated by calculating a central difference \textit{i.e.} the discrete time derivative of the signal, shown in Fig. 1F, which physically correlates to the velocity of the cantilever.
In total, using the presented filters, the SNR for the whole signal can be increased 20-fold from 1.1 to 22.0.

\subsubsection*{Calibration protocol}

Using correct values for the optimization procedure is essential for the success of the localization. 
Parameters, such as $\uptheta_\mathrm{max}$, $\upeta$ and $f_\mathrm{res}$ vary between SMOL devices and physical boundary conditions while $l_0$, $B_r$ and $V$ are fixed.
Hence, the first three parameters have to be determined for each SMOL device at the start of a measurement series with constant environmental conditions.
The resonance frequency $f_\mathrm{res}$ can be found by sweeping the excitation frequency and measuring the resulting frequency of the SMOL signal.
Correctly determining the maximum deflection angle $\uptheta_\mathrm{max}$ is intricate since it correlates to the distance $||\mathbf{r}||^3$ to the sensors (Eq. S9).
One way to resolve this issue is by positioning the SMOL device at the planar (0,0) location with a known $z$-value which leads to $||\mathbf{r}||^3 \approx z^3$.
For small angles of $\uptheta_\mathrm{max}$, $\cos(\uptheta_\mathrm{max}) \approx \mathrm{const.}$ and  $\sin(\uptheta_\mathrm{max}) \sim \uptheta_\mathrm{max}$, meaning that, according to Eq. S10,  $B \sim \uptheta_\mathrm{max}/z^3$.
Therefore, one of both parameters has to be fixed to determine the other.
In the calibration, $\uptheta_\mathrm{max}$ and $\upeta$ are additional optimization parameters and $z$ is fixed to accommodate this correlation and the environment-dependent damping of the cantilever.


\subsubsection*{SMOL device} 

A wireless SMOL device (Fig. 1A) consists two main parts:
a housing and a mechanically resonant structure with an attached magnet.
The housings were designed with a computer-aided design (CAD) software (Inventor Professional 2021, Autodesk, US) and printed using a stereolithography 3D printer (3L, Formlabs, US) with 50 µm resolution and a translucent resin (Clear V4, Formlabs).
The detailed CAD designs of the housings are shown in Fig. S7.
Inside the housing, a cavity was designed to allow the attachment of the cantilever and sufficient space for bending.
The oscillation frequency is tuneable by choice of cantilever dimensions and material to achieve frequencies within the range of approximately 50 Hz to 500 Hz.
For the helical SMOL robot, a $\sim$3.5 mm $\times$ 0.5 mm $\times$ 30 µm stripe (C1095 spring steel, Precision Brand, US) was cut by laser (MPS Advanced, Coherent, US) to form the cantilever, and two $\diameter$ 1 mm $\times$ 0.5 mm cylindrical NdFeB magnets (N52, Guys Magnets, UK) with axial magnetization were attached to the end of the cantilever using cyanoacrylate adhesive (Loctite 401, Henkel, Germany).
The resulting resonance frequency was 181 Hz.
The gradient-pulled millirobot used a $\sim$3.5 mm $\times$ 0.2 mm $\times$ 20 µm stripe steel stripe and a single $\diameter$ 1 mm $\times$ 1 mm magnet with the magnetization axis parallel to the cantilever axis.
The cantilever was tightly fixed to the housing with adhesive (UHU Hart, UHU, Germany) and 30 min curing at 60°C.
The total theoretical magnetic moment of the used magnets amounts to 0.89 mAm². 
The cantilever with fixed magnets was inserted into the housing cavity (or threaded through the opening) and closed with a 3D-printed cover.
Here, a resonance frequency of 103.5 Hz was obtained.
In total, the size of the prototype trackers are $\diameter$ 3 mm $\times$ 7 mm (helical design) and 3 mm $\times$ 3.9 mm $\times$ 2 mm $\times$ (gradient-pulled SMOL design). 

\subsubsection*{Simulation of the SMOL method} 

To simulate the B-field signal of the SMOL device after the excitation, a damped harmonic oscillator model according to Eq. S2 was used with a maximum angular deflection $\uptheta_\mathrm{max} = 17.8$°, a resonance frequency $f_\mathrm{res} = 103.5$ Hz and an linear damping coefficient $\upeta = 1.1$ s$^{-1}$ (Eq. S3). 
These parameters were determined by calibration of the SMOL device for experimental accuracy measurements.
A simulation of an optimized and realistic SMOL device was performed using $\uptheta_\mathrm{max} = 30$°, $f_\mathrm{res} = 160$ Hz, $\upeta = 1.1$ s$^{-1}$ and $N = 6$.
First, additional system dependent variables such as cantilever length ($l_0 = 1.5$ mm), magnetic moment, position and orientation were defined and the movement of the magnetic moment in space was calculated by Eq. S6--S8.
An accurate representation of the real measurement was achieved by calculating the time-dependent Eq. S5 from the cantilever movement at all sensors locations and adding real recorded noise to the signal.
A randomized phase shift of the added noise was introduced to reflect the arbitrary noise phase in the real measurements.
The numerical simulation was implemented by a customized code in MATLAB.

\subsubsection*{Details on the closed-loop control method}

Control of the gradient-pulled millirobot was performed with the setup and control scheme, as shown in Figs. S8A and B.
Initially, the starting position $\mathbf{p}_0$ and orientation $\mathbf{u}_0$ of the robot is determined by randomly activating one of the two excitation coils until sufficient signal strength is reached.
The vector $\mathbf{P}_i$ from the current (denoted by index $i$) position $\mathbf{p}_i$ towards the goal position $\mathbf{p}_{i+1}$ is calculated, which corresponds to the necessary force direction vector $\mathbf{D}_i = \hat{\mathbf{F}_i}$:
\begin{equation}
    \mathbf{D}_i = \hat{\mathbf{P}_i} = \frac{\mathbf{p}_{i+1}-\mathbf{p}_i}{||\mathbf{p}_{i+1}-\mathbf{p}_i||}.
\end{equation}
The coil current $I_\mathrm{i}$ at the current step $i$ is controlled by four voltage-controlled analog outputs of the DAQ
\begin{equation}\label{eq:supp_coil_current}
    I_i = I_\mathrm{min} + (I_\mathrm{max}-I_\mathrm{min})\frac{||\mathbf{p}_{i+1}-\mathbf{p}_i||}{p_\mathrm{thr}},
\end{equation}
where $I_\mathrm{max} = 8$ A is the limiting current, $I_\mathrm{min} = 6$ A is the minimum current and $p_\mathrm{thr} = 3$ mm is the threshold distance at which the maximum current is reduced for fine-maneuvering of the SMOL robot.
A magnetic field look-up table $\mathbf{B}$, simulated with a 0.5 mm mesh resolution using COMSOL (Version 5.6, Comsol Multiphysics, Germany), of each coil $j \in \{1,...,4\}$ at the current location is used to calculate the required linear combination of possible vectors $\mathcal{L}\{\mathbf{B}_{i,j}\}$ to achieve the desired force vector direction.
No torsional forces of the B-field on the magnet are assumed, hence, the B-field direction and magnitude correlate to the force $\mathbf{F}_i$ acting on the magnet. 
The current-ratio $\uplambda_{j,k}$ between a first coil $j$ and second coil $k$ ($k \in \{1,...,4\}$, $k > j$) for planar movement is determined by solving the linear equation system. And the solution is:
\begin{equation}\label{eq:supp_current_factor}
    \uplambda_{j,k} = \frac{-\mathbf{B}_{i,k}\times\mathbf{D}_{i}}{\mathbf{B}_{i,j}\times\mathbf{D}_{i} - \mathbf{B}_{i,k}\times\mathbf{D}_{i}}.
\end{equation}
The scaling factor $\uplambda_{j,k}$ is applied for coil $j$ and $k$ as follows:
\begin{equation}\label{eq:supp_current_j}
    I_j = I_i \frac{1-\uplambda_{j,k}}{\uplambda_{j,k}} \;\;\; \mathrm{if} \;\;\; 0 < \uplambda_{j,k} < 0.5,
\end{equation}
which is the case when coil $j$ needs less current than coil $k$, and
\begin{equation}\label{eq:supp_current_k}
    I_k = I_i \frac{\uplambda_{j,k}}{1-\uplambda_{j,k}} \;\;\; \mathrm{if} \;\;\; 0.5 < \uplambda_{j,k} < 1,
\end{equation}
which is the case when coil $k$ needs less current than coil $j$.
For $\uplambda_{j,k} = 0.5$, both coils are operated with the same current $I_i$.
The boundary cases ($\uplambda_{j,k} = 0$ or $\uplambda_{j,k} = 1$) occur in the hypothetical case when the direction $\mathbf{B}_{i,j}$ of a single coil $j$ is identical to $\mathbf{D}_{i}$.
Then, only coil $j$ is activated with a current of $I_i$ and the calculation of Eqs. S13, S14 and S15 is skipped.
The current is applied for a time $t_\mathrm{act} = 30$ ms, after which the position $\mathbf{p}_i$ is updated by the localization.
Once the criterion for the position difference 
\begin{equation}
    ||\mathbf{p}_{i+1}-\mathbf{p}_i|| \leq 0.8 \ \mathrm{mm}.
\end{equation}
is fulfilled, the next goal position $\mathbf{p}_{i+1}$ is set.
Then, in order to achieve large steering angles (when defining a new goal position), the actuation time $t_\mathrm{act}$ is additionally controlled depending on the angle $\upalpha$ between the current planar orientation $\mathbf{u}_i$ and goal orientation $\mathbf{u}_{i+1}$:
\begin{equation}\label{eq:supp_coil_time}
    t_\mathrm{act} = t_\mathrm{min} + (t_\mathrm{max}-t_\mathrm{min})\frac{|\upalpha|}{180^{\circ}},
\end{equation}
where $t_\mathrm{min} = 30$ ms is the lower limit and $t_\mathrm{max} = 80$ ms is the upper limit actuation time.
Eq. S17 is only applied for sharp turns, $|\upalpha|>30$°, to guarantee the correct alignment of the magnetic moment axis with $\mathbf{D}_i$.
This is crucial for a successful localization after sharp turns, since the SMOL excitation field, contrary to the actuation B-field, must be applied perpendicular to the actuation direction $\mathbf{D}_i$ to excite the SMOL device.

\subsubsection*{Details on localization accuracy}

For spatial and angular accuracy measurements (Fig. 2A-D), a three axis manual linear stage (3x XR50P/M with XR25-XZ, ThorLabs, Germany) with 10 µm resolution and a manual rotation stage (XRR1, ThorLabs) with 0.1° resolution were used to accurately translate and rotate the SMOL device, respectively. 
The reference position was chosen to be approximately (0, 0, 80 mm) with the SMOL devices' main axis pointing in $+y$-direction
and the magnet oscillating in the $x$-$y$-plane. 
Each measurement was independently repeated for 20 times.
The excitation coil was translated when the full excitation of the SMOL device was not achieved.
Mean absolute errors and standard deviations were calculated and compared to the ground truth difference of two respective positions determined by the translation stages. 

\subsubsection*{Endoscope tracking} 

For the demonstration of surgical tool tracking, a flexible endoscope (RIWO D-URS Sensor-Ureteroreno-endoscope 9 Ch NL 600 mm, Richard Wolf, Germany) with a total length of 600 mm and a two-sided 300° range of motion on the flexible end was used.
The static magnetic field created by the endoscope amounts to $\sim$45 µT in close proximity of the tip and up to $\sim$120 µT along the endoscope axis.
A SMOL device was attached to the front tip using adhesive tape.
The {\it{in vitro}} kidney organ was produced with a previously published procedure \cite{adams2017} using a transparent urethane rubber (Clear Flex 30, Smooth-On, US) and it was down-scaled to 80\% of the original size.
Before the tracking process, the kidney model was lubricated with glycerol and the control of the endoscope was done manually.

\subsubsection*{Localization and ultrasound imaging in \textit{ex vivo} brain tissues} 

Porcine brains were obtained from a local butcher, transported on ice and stored in the fridge at 4°C.
All experiments were done within approximately 12 h after the sacrifice of the animal.
The brain sample was hydrated with phosphate-buffered saline solution (PBS, Sigma-Aldrich) and put in a 60 mm $\times$ 50 mm $\times$ 25 mm container for measurements at room temperature.
For US imaging, a handheld US machine (iQ+, Butterfly Network, US) was used with the setting "MSK-Soft Tissue" at a frequency of 1-10 MHz, with a thermal index for soft tissue (TIS) of 0.01 and a mechanical index (MI) of 0.28.
The US probe was set in contact with the biological tissues using US contacting gel (Aquasonic 100, Parker Laboratories, US).
Correlation between SMOL measurements and US images was achieved by referencing the US position to the fixed container walls. 

\subsubsection*{Rotation matrices}

A rotation matrix $\mathbf{R}_i$ is a transformation matrix used to rotate a vector $\mathbf{v}$ around the axis $i$ with angle $\uptheta$ in Euclidean space.
For a unit quaternion $\mathbf{q} = q_0 + q_1 i + q_2 j + q_3 k$ the rotation matrix $\mathbf{R}_q$ is given by
\begin{equation}\label{eq:supp_Rq}
	\mathbf{R}_q(\mathbf{q}) = 2\left( \matrix{ 0.5 - q_2^2 - q_3^2  & q_1 q_2 - q_3 q_0 & q_1 q_3 + q_2 q_0 \cr q_1 q_2 + q_3 q_0 & 0.5 - q_1^2 - q_3^2 & q_2 q_3 - q_1 q_0 \cr  q_1 q_3 - q_2 q_0 & q_2 q_3 + q_1 q_0 & 0.5 - q_1^2 - q_2^2 \cr} \right).
\end{equation}
The rotation matrix around the $y$-axis, $\mathbf{R}_y$, is given by:
\begin{equation}\label{eq:supp_Ry}
	\mathbf{R}_y(\uptheta) = \left( \matrix{ \cos(\uptheta) & 0 & \sin(\uptheta) \cr 0 & 1 & 0 \cr -\sin(\uptheta) & 0 & \cos(\uptheta) \cr} \right).
\end{equation}

\subsubsection*{Coefficient of determination}

For a data set consisting of $k$ points with values $d_{1}$ to $d_{k}$ and fitting values of $f_{1}$ to $f_{k}$, the sum squared error (SSE) is defined by:
\begin{equation}\label{eq:supp_SSE}
	\mathrm{SSE} = \sum^{k}_{i=1} (d_i - f_i)^2.
\end{equation}
The total sum of squares, in contrast, is defined by
\begin{equation}\label{eq:supp_TSS}
	\mathrm{TSS} = \sum^{k}_{i=1} (d_i - \bar{d})^2,
\end{equation}
 with $\bar{d}$ being the mean of the original data set:
\begin{equation}\label{eq:supp_mean}
	\bar{d} = \frac{1}{k}\sum^{k}_{i=1} d_i.
\end{equation}
Finally, the coefficient of determination R$^2$ can be calculated by
\begin{equation}\label{eq:supp_R2}
	\mathrm{R}^2 = 1 - \frac{\mathrm{SSE}}{\mathrm{TSS}}.
\end{equation}
This coefficient is unit-less and typically ranges between 0 and 1, with 1 being a perfect fit of $f_i$ to $d_i$.
Hence, it can be directly used as a measure of quality for data fitting procedures.

\subsubsection*{Statistical analysis}

Taking $d_{i,j}$ as measured values and $f_{i,j}$ as the ground truth value over multiple datasets $j = 1 ... n$, the mean absolute error (MAE), as a measure of accuracy, is defined by
\begin{equation}\label{eq:supp_MAE}
	\mathrm{MAE} = \frac{1}{kn}\sum^{n}_{j=1}\sum^{k}_{i=1} |d_{i,j} - f_{i,j}|.
\end{equation}
For differential measurements over multiple datasets, the mean of the reference dataset $\bar{d}_{\mathrm{ref}}$ (Eq. S22) is subtracted:
\begin{equation}\label{eq:supp_MAE_diff}
	\mathrm{MAE}_{\mathrm{diff}} = \frac{1}{kn}\sum^{n}_{j=1}\sum^{k}_{i=1} |d_{i,j} - \left(f_{i,j} + \bar{d}_{\mathrm{ref}} \right) |.
\end{equation}
The standard deviation $\sigma$ is defined as
\begin{equation}\label{eq:supp_std}
	\sigma = \sqrt{\frac{1}{kn-1}\sum^{n}_{j=1}\sum^{k}_{i=1} \left(\bar{d}_j -d_{i,j} \right)^2}.
\end{equation}
For circular data, \textit{i.e.} when $d_{i,j}$ and $f_{i,j}$ are given in radians, circular statistics were applied ({\it{52}}).
The differential circular MAE is defined by
\begin{equation}\label{eq:supp_MAE_circ}
	\mathrm{MAE}_{\mathrm{diff}}^{\mathrm{circ}} = 
	\mathrm{arg}\left( \sum^{n}_{j=1}\sum^{k}_{i=1} \exp\left(\mathrm{i}|d_{i,j} - (f_{i,j} + \bar{d}_\mathrm{ref})|\right) \right)
\end{equation}
and the standard deviation by
\begin{equation}\label{eq:supp_std_circ}
		\sigma^{\mathrm{circ}} = \sqrt{\left(-2\ln{R}\right)}
\end{equation}
with
\begin{equation}\label{eq:supp_std_circ_R}
		R = |\sum^{n}_{j=1}\sum^{k}_{i=1} \exp{\left(\mathrm{i}(\bar{d}_j -d_{i,j}) \right)}|
\end{equation}
and circular mean
\begin{equation}\label{eq:supp_mean_circ}
		\bar{d} = \mathrm{arg}\left( \sum^{k}_{i=1} \exp{\left(\mathrm{i} d_i \right)}\right)
\end{equation}

\section*{References} 
\begin{itemize}
    \item [51.] Chow.~T. {\it{Introduction to Electromagnetic Theory: A Modern Perspective}} (Jones \& Bartlett Learning, 2006).
    \item [52.] Fisher~N.~I. {\it{Statistical Analysis Of Circular Data}} (Cambridge Univ.~Press, 1995).
\end{itemize}
\end{document}